\documentclass[AMA,STIX2COL,final]{MRM}
\articletype{Full Paper}%
\received{28 Jul 2022}
\revised{19 Dec 2022}
\accepted{8 Feb 2023}

\renewcommand{\bf}[1]{\textbf{#1:}}
\usepackage{etoolbox}
\newtoggle{SUPMATERIAL}
\toggletrue{SUPMATERIAL}

\usepackage{ifdraft}
\usepackage{color}
\renewcommand{\r}{\textcolor{red}}
\ifoptionfinal{\newcommand{\delete}[1]{}}
{
\newcommand{\stkout}[1]{\ifmmode\text{\sout{\ensuremath{#1}}}\else\sout{#1}\fi}
\newcommand{\delete}[1]{\r{\stkout{#1}}}
}

\usepackage{marginnote}
\newcommand{\R}[1]{\ifoptionfinal{}{\marginnote{\textcolor{red}{#1}}}}
\renewcommand{\b}{\ifoptionfinal{}{\textcolor{blue}}}

\usepackage{bbold}
\usepackage{graphicx}
\usepackage{pdfpages}
\usepackage{adjustbox}
\usepackage{booktabs}
\usepackage{makecell}

\usepackage[capitalise, noabbrev]{cleveref}

\usepackage{subcaption}
\newcommand{\labelphantom}[1]{%
  \parbox{0pt}{\phantomsubcaption\label{#1}}%
}
\graphicspath{{./Figures/}}

\definecolor{gray}{RGB}{220,220,220}

\title{Bayesian MRI Reconstruction with Joint Uncertainty Estimation using Diffusion Models\protect\thanks{Parts of this work were presented at the ISMRM 2022.}}
\newcommand{\authorA}{Guanxiong Luo}
\newcommand{\authorC}{Martin Heide}
\newcommand{\authorB}{Moritz Blumenthal}
\newcommand{\authorD}{Martin Uecker}

\newcommand{\affilA}{Institute of Biomedical Imaging, Graz University of Technology, Graz, Austria}
\newcommand{\affilB}{Institute for Diagnostic and Interventional Radiology of the University Medical Center G\"ottingen, Germany}
\newcommand{\affilC}{German Centre for Cardiovascular Research (DZHK), Partner Site G\"ottingen, Germany}
\newcommand{\affilD}{Cluster of Excellence ``Multiscale Bioimaging: from Molecular Machines to Networks of Excitable Cells'' (MBExC), University of G\"ottingen, Germany}    

\newcommand{\corAdress}{Guanxiong Luo, University Medical Center 
Göttingen, Institute for Diagnostic and Interventional Radiology, 
Robert-Koch-Str. 40, 37075 Göttingen, Germany.}

\newcommand{\corMail}{Email: guanxiong.luo@med.uni-goettingen.de}

\newcommand{\Funding}{
    We acknowledge funding by the ``Niedersächsisches Vorab" funding line of the Volkswagen Foundation.
}

\author{\authorA$^2$}{\orcid{0000-0001-8005-4639}}
\author{\authorB$^{1,2}$}{\orcid{0000-0002-2127-8365}}
\author{\authorC$^{2}$}{\orcid{0000-0002-4129-7395}}
\author{\authorD$^{1,2,3,4}$}{\orcid{0000-0002-8850-809X}}
\authormark{Luo \textsc{et al}}

\address[1]{\affilA}
\address[2]{\affilB}
\address[3]{\affilC}
\address[4]{\affilD}

\corres{\corAdress\ \corMail}

\presentaddress{\corAdress}

\finfo{\Funding}
\ifoptionfinal{}{\details{Word count: 5266, Figures: 10, Tables: 0.}}

\jnlcitation{\cname{%
   		\author{Luo, G.},
   		\author{Heide, M.},
   		\author{Blumenthal, M.}, and 
   		\author{Uecker, M.}}, \cyear{2022}.
   		\ctitle{MRI Reconstruction via Data-Driven Markov
   		Chains with Joint Uncertainty Estimation}. \cjournal{Magn. Reson. Med.} \cvol{2022;xx:xx--xx}.}

\begin{document}
\abstract{\section{Purpose} We introduce a framework that enables efficient sampling from learned probability distributions for MRI reconstruction.
	\section{Method} Samples are drawn from the posterior distribution given the measured k-space using the Markov chain Monte Carlo (MCMC) method, different from conventional deep learning-based MRI reconstruction techniques. In addition to the maximum a posteriori (MAP) estimate for the image, which can be obtained by maximizing the log-likelihood indirectly or directly, the minimum mean square error (MMSE) estimate and uncertainty maps can also be computed from those drawn samples. The data-driven Markov chains are constructed with the score-based generative model learned from a given image database and are independent of the forward operator that is used to model the k-space measurement. 	\section{Results} We numerically investigate the framework from these perspectives: 1) the interpretation of the uncertainty of the image reconstructed from undersampled k-space;
	2) the effect of the number of noise scales used to train the generative models;
	3) using a burn-in phase in MCMC sampling to reduce computation;
	4) the comparison to conventional $\ell_1$-wavelet regularized reconstruction;
	5) the transferability of learned information; and 
	6) the comparison to fastMRI challenge.
	\section{Conclusion} A framework is described that connects the diffusion process and advanced generative models with Markov chains. We demonstrate its flexibility in terms of contrasts and sampling patterns using advanced generative priors and the benefits of also quantifying the uncertainty for every pixel.}

\maketitle

\section{Introduction}
Modern Magnetic Resonance Imaging (MRI) formulates reconstruction from raw data in Fourier space (k-space) as an inverse problem. 
Undersampling to reduce acquisition time then leads to an ill-posed reconstruction problem.
To solve this problem, parallel imaging can exploit spatial information from multiple receive coils in an
extended forward model \cite{Pruessmann_Magn.Reson.Med._2001}.
Compressed sensing uses the sparsity of images in a transform domain (i.e. wavelet domain, finite differences) as prior knowledge.
Combined with incoherent sampling this allows recovery of sparse images from highly
undersampled data \cite{Lustig_Magn.Reson.Med._2007,Block_Magn.Reson.Med._2007}.
Learning-based techniques for compressed sensing include methods
using dictionary learning \cite{Ravishankar_IEEETrans.Med.Imag._2010} or a patch-based nonlocal operator \cite{Qu_Med.ImageAnal._2014}.

\begin{figure*}
		\centerline{\includegraphics[width=0.8\textwidth]{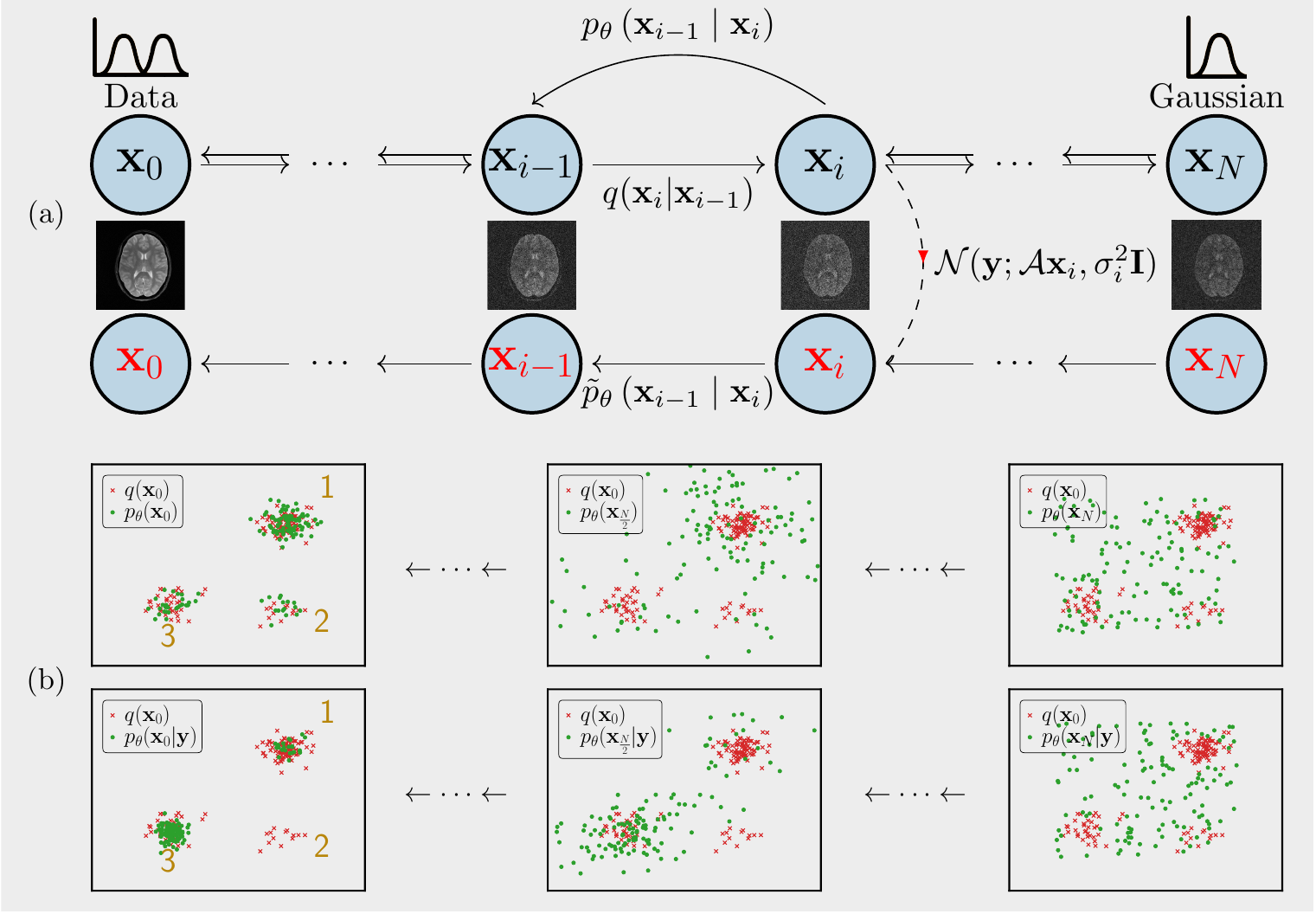}}
	\caption{Overview of the proposed method. (a) The unknown data distribution $q(\mathbf{x}_0)$ of the training images goes through repeated Gaussian diffusion and finally reaches a known Gaussian distribution $q(\mathbf{x}_N)$, and this process is reversed by learned transition kernels $p_{\mathbf{\theta}}\left(\mathbf{x}_{i-1} \mid \mathbf{x}_{i}\right)$. To compute the posterior of the image $p(\mathbf{x}|\mathbf{y})$, a new Markov chain $\tilde{p}_{\mathbf{\theta}}\left(\mathbf{x}_{i-1} \mid \mathbf{x}_{i}\right)$ is constructed by incorporating the measurement model into the reverse process (red chain). (b) Training samples (red dots) from a mixture of bivariate Gaussian distribution are shown. The upper and bottom rows illustrate how samples (green dots) gradually gather around training samples in the reverse process, without and with the observation, respectively. In this example, the likelihood for the observation was a bivariate Gaussian mixture, so that cluster 2 has a lower and cluster 3 has a higher probability.}
	\label{fig:overview}
\end{figure*}
In recent years, the application of deep learning
pushed these ideas forward by integrating learned prior knowledge \cite{Wang_I.S.Biomed.Imaging_2016}.
Most of these methods can be classified into two categories: First, methods that
unroll the existing iterative reconstruction algorithms into a neural network
and train their parameters by maximizing the similarity to a ground truth.
In Ref. [\citen{Yang_NIPS_2016}], the authors replaced the handcrafted regularization term
with convolution layers, and derived a neural network from the iterative
procedure of the Alternating Direction Method of Multipliers (ADMM) algorithm.
Ref. [\citen{Aggarwal_IEEETrans.Med.Imag._2019,Hammernik_Magn.Reson.Med._2017}] investigated similar approaches.
The downside of this kind of method is the need for supervised training, which
requires raw k-space data with fixed known sampling patterns and corresponding
ground truth images.
The second category consists of methods that learn a prior from high-quality images,
then plug it into existing iterative algorithms as a regularization term.
In Ref. [\citen{Tezcan_IEEETrans.Med.Imag._2018,Luo_Magn.Reson.Med._2020, Liu_Magn.Reson.Med._2020}], the image prior was constructed
with a variational auto-encoder \cite{Kingma_ICLR_2014}, a
denoising auto-encoder \cite{Alain_J.Mach.Learn.Res._2014} and an autoregressive 
generative model \cite{Salimans_ICLR_2017}, respectively.
These methods then compute a maximum a posterior (MAP) as the estimator 
of the image.
These types of methods separate the learned information from the encoding
matrix (sampling pattern in k-space and coil sensitivities), which permits
more flexibility in practice because they allow the acquisition patterns 
and receive coils to change without retraining.
Generative adversarial networks were also used for image reconstruction 
in Ref. [\citen{Mardani_IEEETrans.Med.Imag._2018}].
There, the discriminator is used to confine the space of the output of a generator that is designed 
to generate images with conformity to k-space data. 

Although deep learning based approaches provide promising results,
worries about the uncertainty caused by undersampling strategies and algorithms have limited their usage in clinical practice until now.
Therefore,  the uncertainty assessment constitutes an important step for deep learning based approaches. 
The uncertainty is two-fold: 1) the uncertainty of weights inside the
neural network \cite{Blundell_ICML_2015,Narnhofer_IEEETrans.Med.Imag._2022}; and
2) the uncertainty introduced by the missing k-space data points.
The uncertainty from missing k-space data points can be addressed in a Bayesian imaging framework. We refer the readers to Ref. [\citen{Calvetti_Inverse.Prob._2008,Stuart_Acta.Numerica_2010}].
In Ref. [\citen{Luo_Magn.Reson.Med._2020}], the MAP estimator is used,
but it provides only the mode of the posterior density $p(\mathbf{x}|\mathbf{y})$
and practical optimization may also even only provide a local maximum.
In the setting of Bayesian inference, it is possible to investigate the full shape of posterior distribution $p(\mathbf{x}|\mathbf{y})$.
In particular, it is possible to draw sample from the posterior distribution
for priors based on diffusion models
using the Markov chain Monte Carlo (MCMC) method
as described previously by Jalal et al. \cite{Jalal_NIPS_2021}
and others \cite{Luo__2022,Chung_Med.Image.Anal._2022, Levac_arXiv_2022},
\b{which are closely related to the present work. Jalel et al. use Langevin sampling to
sample the posterior using score-based generative model and this is extendend
in Ref. [\citen{Levac_arXiv_2022}] to also include a motion model. The method 
in Ref. [\citen{Chung_Med.Image.Anal._2022}] uses the predictor-and-corrector
framework proposed in Ref. [\citen{Song_ICLR_2021}]. These publications
point out the relationship to Bayesian reconstruction and show some results
related to uncertainty estimation, but a complete Bayesian formulation
of this framework applied to MRI multi-channel reconstruction is not
provided. A general problem with this approch is the large number of
iterations required during sampling, e.g. Ref. [\citen{Chung_Med.Image.Anal._2022}]
reports the use of several thousands of iterations.}\R{R3.1}

Following these ideas, a generic framework for MRI reconstruction emerges, which is based on a series of publications related to generative models \cite{Song_ICLR_2021,Hyvarinen_J.Mach.Learn.Res._2005,Vincent_NeuralComput._2011,Sohldickstein_ICML_2015,Song_NIPS_2019}, in which the essential idea is to:
1) systematically and slowly destroy the underlying prior knowledge in a data distribution through an iterative forward diffusion process;
2) learn a reverse diffusion process that restores the patterns by a so-called score-based neural network
and 3) incorporate the forward model of the measurement into the learned reverse process. 
The general picture of the proposed method is illustrated in \cref{fig:overview}.

In the present work, 
we recapitulate the framework of Bayesian reconstruction and
score-based diffusion models and
numerically investigate this framework from the 
following different perspectives:
1) the interpretation of the uncertainty of the image reconstructed from undersampled k-space;
2) the effect of the number of noise scales used the generative models \b{on image quality on computation time};
3) \delete{the effect of} using a burn-in phase in MCMC sampling \b{to reduce computation};
4) the comparison to conventional $\ell_1$-wavelet regularized reconstruction;
5) the transferability of learned information;  and 6) the comparison to fastMRI challenge\cite{Zbontar_arXiv_2019,Zaccharie_arXiv_2020}.

\section{Theory}

\subsection{Magnetic Resonance Image Reconstruction as Bayesian Inference}

We consider image reconstruction as a Bayesian problem
where the posterior of image $p(\mathbf{x}|\mathbf{y})$ 
given with the measured data $\mathbf{y}$ and
a prior $p(\mathbf{x})$ learned from a database of images \cite{Luo_Magn.Reson.Med._2020, Calvetti_Inverse.Prob._2008, Stuart_Acta.Numerica_2010}.
Here, the image is denoted as $\mathbf{x}\in \mathbb{C}^{n\times n}$, where $n\times
n$ is the size of image, and $\mathbf{y} \in \mathbb{C}^{m \times m_C}$ is the
vector of $m$ complex-valued k-space samples from $m_C$ receive coils.
Assuming the noise $\eta$ circularly-symmetric normal with zero mean and
covariance matrix $\sigma^2_{\eta} \mathbf{I}$, the
likelihood $p(\mathbf{y}|\mathbf{x})$ for observing the $\mathbf{y}$ 
determined by $\mathbf{y} = \mathcal{A} \mathbf{x}+\eta$
and given the image $\mathbf{x}$ is given by a complex normal distributions
\begin{align}
	p(\mathbf{y}|\mathbf{x}) & = \mathcal{CN}(\mathbf{y}; \mathcal{A}\mathbf{x}, \sigma^2_{\eta} \mathbf{I}) \nonumber\\
	& = (\sigma_{\eta}^2\pi)^{-N_p} e^{\text{-}\|\sigma_{\eta}^{-1} \cdot (\mathbf{y} - \mathcal{A}\mathbf{x})\|_2^2}~,
	\label{eq:forw} 
\end{align}

where $\mathbf{I}$ is the identity matrix, $\sigma_{\eta}$ the standard deviation of the noise, $\mathcal{A}\mathbf{x}$ is the mean and $N_p$ is the length of the k-space data vector.
$\mathcal{A}:\mathbb{C}^{n \times n}\rightarrow\mathbb{C}^{m \times m_C}$ is the forward operator and given by $\mathcal{A}=\mathcal{PFS}$,
where $\mathcal{S}$ are the coil sensitivity maps,
$\mathcal{F}$ the two-dimensional Fourier transform, and
$\mathcal{P}$ the
k-space sampling operator.
According to Bayes' theorem the posterior density function
$p(\mathbf{x}|\mathbf{y})$ is then
\begin{align}
	p(\mathbf{x}|\mathbf{y}) &= \frac{p(\mathbf{y}|\mathbf{x})\cdot p(\mathbf{x})}{p(\mathbf{y})} ~.
	\label{eq:bayes}
\end{align}
In this work, the reconstruction is based on the sampling of this posterior
distribution. We utilize an efficient technique based on
the Markov Chain Monte Carlo method with the application of a diffusion probabilistic 
generative model. This consists of two processes:
1) a forward diffusion process which converts a complicated
distribution used as prior for the image into a simple Gaussian distribution; and
2) a learned finite-time reversal of this diffusion process
with which a Gaussian distribution is gradually transformed back to 
the posterior (cf. \cref{fig:overview}).

\subsection{The Forward Diffusion Process}
In probabilistic diffusion models, the data distribution characterized by
density $q(\mathbf{x}_{0})$ is gradually converted into an analytically
tractable distribution (Gaussian noise) \cite{Sohldickstein_ICML_2015}.
The image $\mathbf{x}_{0}$ is
perturbed with a sequence of noise scales $0 = \sigma_0 < \sigma_1 < \cdots < \sigma_N$. 
When the number of steps used for discretization $N \rightarrow \infty$, the diffusion process becomes a
continuous process.
Here, we consider the discrete Markov chain
\begin{equation}
	\mathbf{x}_{i}=\mathbf{x}_{i-1} + \mathbf{z}_{i-1}, \quad i=1, \cdots, N,
\end{equation}
where $\mathbf{z}_{i-1} \sim \mathcal{C N}(\mathbf{0}, (\sigma_{i}^{2}-\sigma_{i-1}^{2}) \mathbf{I})$,
i.e. the $i$-th transition kernel is then given by
\begin{equation}
   q(\mathbf{x}_{i}|\mathbf{x}_{i-1}) = \mathcal{CN}(\mathbf{x}_{i}; \mathbf{x}_{i-1}, (\sigma_{i}^2-\sigma_{i-1}^2)\mathbf{I})~.
   \label{eq:markov}
\end{equation}
Instead of doing transitions step by step \cite{Song_ICLR_2021, sarkka2019applied}
a single
perturbation kernel
\begin{equation}
q(\mathbf{x}_i\mid \mathbf{x}_0) = \mathcal{CN}\left(\mathbf{x}_i ; \mathbf{x}_0, \sigma^{2}_i \mathbf{I}\right)
\label{eq:5}
\end{equation}
can be computed
as a convolution of Gaussians. With Bayes’ theorem we can write:
\begin{equation}
     q\left(\mathbf{x}_{i-1} \mid \mathbf{x}_{i}, \mathbf{x}_{0}\right) =
q\left(\mathbf{x}_{i} \mid \mathbf{x}_{i-1}\right) 
\frac{q\left(\mathbf{x}_{i-1} \mid \mathbf{x}_{0}\right)}{q\left(\mathbf{x}_{i} \mid \mathbf{x}_{0}\right)}~.
\end{equation}
Given the initial image $\mathbf{x}_0$, the posterior of a single step of the forward process
is then given by (see Appendix \ref{appendix.a})
\begin{equation}
	q(\mathbf{x}_{i-1} \mid \mathbf{x}_{i}, \mathbf{x}_{0}) = \mathcal{CN}\bigl(\mathbf{x}_{i-1} ; \frac{\sigma_{i-1}^{2}}{\sigma_{i}^{2}} \mathbf{x}_{i}+\bigl(1-\frac{\sigma_{i-1}^{2}}{\sigma_{i}^{2}}\bigr) \mathbf{x}_{0}, \tau_i^2 \mathbf{I}\bigr)
	\label{eq:forward-post}
\end{equation}
with variance $\tau_{i}^2 := \left(\sigma_{i}^{2}-\sigma_{i-1}^{2}\right) \left( \sigma_{i-1}^{2} / \sigma_{i}^{2} \right)$.

\subsection{Learning the Reverse Process}

The joint distribution of the reversal diffusion process is characterized by the probability density
\begin{align}
	p(\mathbf{x}_N, \mathbf{x}_{N-1}, \cdots, \mathbf{x}_0) & = p(\mathbf{x}_{N})\prod_{i=1}^{N}p(\mathbf{x}_{i-1}|\mathbf{x}_{i})~,
\end{align}
where $p(\mathbf{x}_{N})$ is the initial Gaussian distribution.
The reverse is given by Kolmogorov's backward equation which has the same form as the forward process \cite{Sohldickstein_ICML_2015, sarkka2019applied}.
Thus, the transitions $p(\mathbf{x}_{i-1}|\mathbf{x}_{i})$ of the reverse process can be parameterized with the
Gaussian transition kernel
\begin{align}
	p\left(\mathbf{x}_{i-1} \mid \mathbf{x}_{i}\right)&=\mathcal{CN}\left(\mathbf{x}_{i-1} ; \boldsymbol{\mu}\left(\mathbf{x}_{i}, i\right), \tau_{i}^{2} \mathbf{I}\right),\label{eq:learnt-tran}
\end{align}
where $\boldsymbol{\mu}\left(\mathbf{x}_{i}, i\right)$
and $\tau_{i}^{2} \mathbf{I}$ are the mean and variance of 
the reverse transitions, respectively.
Here, we learn the mean $\mu_{\mathbf{\theta}}$ of the reverse transitions using a neural network
parameterized by training parameters $\theta$.
Since the learned reverse transitions $p_{\theta}(\mathbf{x}_{i-1}|\mathbf{x}_{i})$ lead
to a new density $p_{\mathbf{\theta}}(\mathbf{x}_0)$, which should  match $q(\mathbf{x}_0)$,
they can be learned by minimizing the cross entropy %
\begin{equation}
	H(p_{\mathbf{\theta}}, q) = -\mathbb{E}_{q(\mathbf{x}_0)}\left[\log p_{\mathbf{\theta}}(\mathbf{x}_0)\right]~.
\end{equation}
Following Ref. [\citen{Sohldickstein_ICML_2015}]
a lower bound $\ell$ can be written in terms of KL divergence between the transition kernel \cref{eq:learnt-tran}
and the posterior of forward process \cref{eq:forward-post}
\begin{align}
	\ell
	= &\sum_{i=2}^{N} \mathbb{E}_{q(\mathbf{x}_0)} \mathbb{E}_{q(\mathbf{x}_i \mid \mathbf{x}_0)} \bigl[
		D_{\mathrm{KL}}(q(\mathbf{x}_{i-1} \mid \mathbf{x}_{i}, \mathbf{x}_{0}) \| \ p_{\boldsymbol{\theta}}(\mathbf{x}_{i-1} \mid \mathbf{x}_{i}))\bigr] \nonumber\\
	= &\sum_{i=2}^{N} \mathbb{E}_{q(\mathbf{x}_0)} \mathbb{E}_{q(\mathbf{x}_i \mid \mathbf{x}_0)}\nonumber\\
	&\Bigl[\frac{1}{\tau_{i}^{2}}\Bigl\|\frac{\sigma_{i-1}^{2}}{\sigma_{i}^{2}} \mathbf{x}_{i}+\bigl(1-\frac{\sigma_{i-1}^{2}}{\sigma_{i}^{2}}\bigr) \mathbf{x}_{0}-\boldsymbol{\mu}_{\boldsymbol{\theta}}(\mathbf{x}_{i}, i)\Bigr\|_{2}^{2}\Bigr]+C,
	\label{eq:kl}
\end{align}

where $C$ is a constant. The derivation of KL divergence between two Gaussian distributions is detailed in Appendix \ref{appendix.f}. Using \cref{eq:5} we can express $\mathbf{x}_i=\mathbf{x}_0 + \mathbf{z}$ with $\mathbf{z} \sim \mathcal{CN}\left(\mathbf{0}, \sigma^{2}_i \mathbf{I}\right)$, 
and obtain
\begin{equation}
	\ell =\sum_{i=2}^{N}\mathbb{E}_{\mathbf{x}_{0}, \mathbf{z}}\left[\frac{1}{\tau_{i}^{2}}\left\|\frac{{\sigma}_{i-1}^{2}}{\sigma_{i}^2} \mathbf{z} + \mathbf{x}_0 -\boldsymbol{\mu}_{\boldsymbol{\theta}}\left(\mathbf{x}_{i}, i\right)\right\|_{2}^{2}\right]+C.
		\label{eq:kl1}
\end{equation}
Thus, we can learn the mean of the reverse transitions by learning to denoise the training data disturbed by noise.
In Ref. [\citen{Hyvarinen_J.Mach.Learn.Res._2005, Vincent_NeuralComput._2011}], the generative model is estimated by minimizing the expected squared distance between the gradient of the log-probability given by the score network and the gradient of the log-probability of the observed data. 
This technique was extended and generalized in Ref. [\citen{Song_NIPS_2019,Song_ICLR_2021}].
In the following, we quickly point out the connection to score matching networks. Let:
\begin{equation}
	\boldsymbol{\mu}_{\boldsymbol{\theta}}\left(\mathbf{x}_{i}, i\right)-\mathbf{x}_{0} = \sigma_{i-1}^{2} \mathbf{s}_{\boldsymbol{\theta}}\left(\mathbf{x}_{i}, i\right),
	\label{eq:mean_s}
\end{equation}
where $\mathbf{s}_{\boldsymbol{\theta}}(\mathbf{x}_i, i)$ denotes the denoising score matching network that is conditional on the index of noise scales $i$. Then, we have
\begin{equation}
	\ell =\sum_{i=2}^{N}\mathbb{E}_{\mathbf{x}_{0}, \mathbf{z}}\left[\frac{\sigma_{i-1}^2}{\tau_{i}^{2}}\left\|\frac{\mathbf{z}}{\sigma_{i}^2}-\boldsymbol{s}_{\boldsymbol{\theta}}\left(\mathbf{x}_{i}, i\right)\right\|_{2}^{2}\right]+C ~.
	\label{eq:kl2}
\end{equation}
Expressing the noise again as $\mathbf{z} = \mathbf{x}_i - \mathbf{x}_0$, we can rewrite
\begin{align}
	 & \mathbb{E}_{\mathbf{x}_0, \mathbf{z}}\left[\left\|\frac{\mathbf{x}_i-\mathbf{x}_0}{\sigma_i^2} - \mathbf{s}_{\boldsymbol{\theta}}(\mathbf{x}_i, i)\right\|_{2}^{2}\right] \nonumber\\
	 & = \mathbb{E}_{q(\mathbf{x}_0)} \mathbb{E}_{q(\mathbf{x}_i \mid \mathbf{x}_0)}\left[\left\|\nabla_{\mathbf{x}_i} \log q(\mathbf{x}_i \mid \mathbf{x}_0) - \mathbf{s}_{\boldsymbol{\theta}}(\mathbf{x}_i, i)\right\|_{2}^{2}\right] 
	\label{eq:score}
\end{align}
which shows that \cref{eq:kl2} is equivalent to score matching. For the later use of the transition kernel, \cref{eq:mean_s} is equivalent to
\begin{equation}
	\boldsymbol{\mu}_{\boldsymbol{\theta}}\left(\mathbf{x}_{i}, i\right)-\mathbf{x}_{i} = \left(\sigma_{i}^{2}-\sigma_{i-1}^{2}\right) \mathbf{s}_{\boldsymbol{\theta}}\left(\mathbf{x}_{i}, i\right) ~.
	\label{eq:mean}
\end{equation}
In summary, the score network is trained via \cref{eq:score} to output the
gradient fields that are used to construct the Markov transitions (\cref{eq:learnt-tran})
which nudges coarse samples $\mathbf{x}_i$ toward finer ones $\mathbf{x}_{i-1}$, namely the reverse process.
In later sections, we will discuss how we construct and train the score networks.

\subsection{Computing the Posterior for MRI Reconstruction}

In order to compute the posterior probability $p(\mathbf{x}|\mathbf{y})$ for the image $\mathbf{x}$ given the
data $\mathbf{y}$, we need to modify the learned reverse process. We achieve this by multiplying each
of the intermediate distributions $p(\mathbf{x}_i)$ with the likelihood term $p(\mathbf{y}|\mathbf{x}_i)$
according to Bayes' theorem.
We use $\tilde{p}\left(\mathbf{x}_{i}\right) = p(\mathbf{x}_i| \mathbf{y})$ to denote the resulting sequence of
intermediate distributions
\begin{equation}
	\tilde{p}\left(\mathbf{x}_{i}\right)\propto p\left(\mathbf{x}_{i}\right) p(\mathbf{y}|\mathbf{x}_i)
\end{equation}
up to the unknown normalization constant.
Following Ref. [\citen{Sohldickstein_ICML_2015}], the transition from $\mathbf{x}_{i+1}$ to $\mathbf{x}_i$ of the modified reverse process is
\begin{equation}
	\tilde{p}\left(\mathbf{x}_{i} \mid \mathbf{x}_{i+1}\right)\propto p\left(\mathbf{x}_{i} \mid \mathbf{x}_{i+1}\right) p(\mathbf{y}|\mathbf{x}_i)~.
	\label{eq:pert}
\end{equation}
The sampling at each intermediate distribution of Markov transitions \cref{eq:pert} is performed with the unadjusted Langevin 
algorithm
\cite{douc2018markov}
\begin{equation}
  \mathbf{x}_i^{k+1} \leftarrow \mathbf{x}_i^{k} + \frac{\gamma}{2}\nabla_{\mathbf{x}_i}\log \tilde{p}(\mathbf{x}_{i}^{k}\mid\mathbf{x}_{i+1}) + \sqrt{\gamma}\mathbf{z},
  \label{eq:ld}
\end{equation}
where $\mathbf{z}$ is standard complex Gaussian noise $\mathcal{CN}(0, \mathbf{I})$.
We now go over to the modified learned process $\tilde{p}_{\boldsymbol{\theta}}(\mathbf{x}_{i}\mid\mathbf{x}_{i+1})$ 
parameterized by $\theta$ and obtain the log-derivative
with respect to $\mathbf{x}_i$ using the learned reverse transitions $p_{\boldsymbol{\theta}}\left(\mathbf{x}_i \mid \mathbf{x}_{i+1}\right)$
as
\begin{equation}
	\nabla_{\mathbf{x}_i}\log \tilde{p}_{\boldsymbol{\theta}}(\mathbf{x}_{i}\mid\mathbf{x}_{i+1}) = \nabla_{\mathbf{x}_i}\log p_{\boldsymbol{\theta}}\left(\mathbf{x}_i \mid \mathbf{x}_{i+1}\right) + \nabla_{\mathbf{x}_i}\log p(\mathbf{y}|\mathbf{x}_i).
\end{equation}
From \cref{eq:learnt-tran} and \cref{eq:mean}, we have
\begin{align}
	\nabla_{\mathbf{x}_i}\log p_{\boldsymbol{\theta}}\left(\mathbf{x}_i \mid \mathbf{x}_{i+1}\right) = \frac{1}{\tau_{i+1}^2}\left(\sigma_{i+1}^{2}-\sigma_{i}^{2}\right) \mathbf{s}_{\boldsymbol{\theta}}\left(\mathbf{x}_{i+1}, i\right),
\end{align}
and from \cref{eq:forw} we have
\begin{align}
	\nabla_{\mathbf{x}_i}\log p(\mathbf{y}|\mathbf{x}_i) & = -\frac{1}{\sigma_\eta^2}(\mathcal{A}^{H} \mathcal{A}\mathbf{x}_i-\mathcal{A}^{H}\mathbf{y}) ~.
\end{align}
After inserting these expressions into \cref{eq:ld} we obtain
\begin{align}
	\mathbf{x}_i^{k+1} \leftarrow \mathbf{x}_i^{k} & + \frac{\gamma}{2\tau_{i+1}^2}(\sigma_{i+1}^{2}-\sigma_{i}^{2}) \mathbf{s}_{\boldsymbol{\theta}}(\mathbf{x}_{i}^k, i) \nonumber\\
	 &-\frac{\gamma}{2\sigma_\eta^2} (\mathcal{A}^{H} \mathcal{A}\mathbf{x}_i^k-\mathcal{A}^{H}\mathbf{y}) + \sqrt{\gamma}\mathbf{z}~.
	\label{eq:ld2}
\end{align}
The starting point for each chain $\mathbf{x}_i^0=\mathbf{x}_{i+1}^\mathrm{K}$
is the last sample from the previous distribution $\tilde{p}(\mathbf{x}_{i+1} \mid \mathbf{x}_{i+2})$ 
after $\mathrm{K}$ Langevin steps.
We found it advantageous to modify the likelihood term
in each step according $\sigma_\eta^2=\tau_{i+1}/\lambda$, which
should approach the variance of the data noise in the last step.
Since the noise variance was unknown for the data set we used, we empirically selected a $\lambda$ that determines how strong the k-space data consistency is relative to the prior. We set $\gamma$ to $2\tau_{i+1}^2$. At last, the algorithm used to sampling the posterior is presented in \cref{alg:seq}.

\begin{algorithm}
\begin{algorithmic}[1]
\State Give the acquired k-space $\mathbf{y}$.
\State Construct the forward operator $\mathcal{A}$ with sampling pattern $\mathcal{P}$ and coil sensitivities $\mathcal{S}$.
\State Set the Langevin steps $\mathrm{K}$, the factor $\lambda$, the start noise level index $\mathrm{N}$, and $\gamma$.
\State Generate $\mathbf{x}_\mathrm{N}^0$ from a suitable Gaussian distribution (e.g., $\mathcal{CN}\sim(0,\mathbf{I})$) \delete{with the zero mean and $\sigma_N^2$ variance}. 
\For{$i$ in \delete{$\{\mathrm{N-1},\cdots,0\}$} \b{$\{\mathrm{N-1},\cdots,1\}$}}
\item  Draw samples from $\tilde{p}(\mathbf{x}_{i}|\mathbf{x}_{i+1})$ by running $\mathrm{K}$ Langevin steps with \cref{eq:ld2}.
\EndFor
\end{algorithmic}
\caption{Sampling the posterior with a Markov chain Monte Carlo method}
\label{alg:seq}
\end{algorithm}

\begin{figure*}
	
		\centering
		\includegraphics[width=0.8\textwidth]{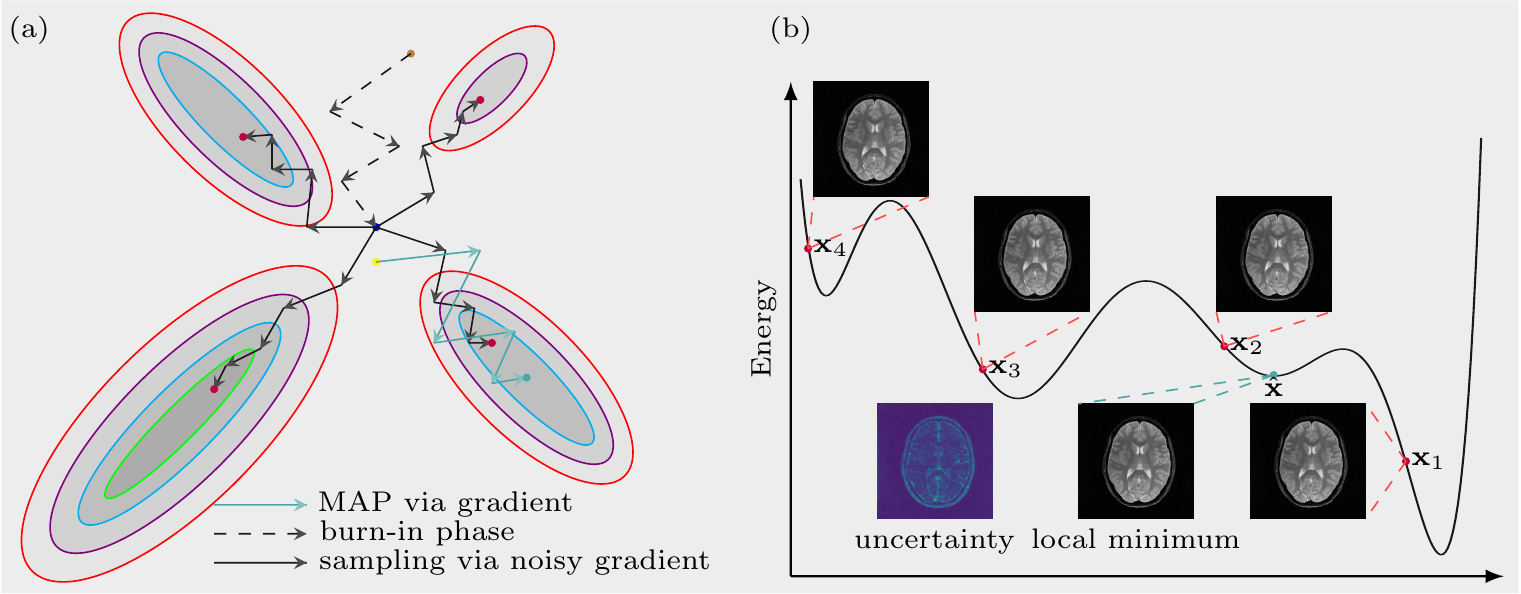}
	
		\caption{Illustration for the sampling of the posterior $p(\mathbf{x}|\mathbf{y})$. (a) The four possible sampling trajectories are indicated the solid lines, sharing the same burn-in phase (dashed line). The MAP approach via gradient descent reaches a locally optimal solution. (b) Possible reconstructions are showed over the energy curve and the uncertainty map is the pixelwise variance over samples. }
		\label{fig:mcmc}
\end{figure*}
 
To characterize the shape of a posterior, we run multiple chains to draw samples in parallel. To reduce the amount of computation, the burn-in phase is introduced as shown in \cref{fig:mcmc}. That means only one chain proceeds through the several beginning noise levels, and after that we
split it up into multiple Markov chains using the sample
from the burn-in phase as initial point indicated by the blue dot. To further reduce computation, we introduce the continuously decreasing noise scales, which reduces the number of iterations when performing Langevin dynamics at each intermediate distribution.
\subsection{The Analysis of Samples}

Given a posterior probability distribution $p(\mathbf{x}|\mathbf{y})$
the minimum mean square error (MMSE) estimator minimizes the mean square error:
\begin{equation}
	{\mathbf{x}}_\mathrm{MMSE} = \arg \min_{\tilde{\mathbf{x}}} \int \|\tilde{\mathbf{x}} - \mathbf{x}\|^2 p(\mathbf{x}|\mathbf{y})d\mathbf{x} = \mathbb{E}[x|y]~.
\end{equation}
The MMSE estimator cannot be computed in a closed form, and numerical approximations are typically required. Since we demonstrated how to generate samples from the posterior in previous sections, let us consider the samples $\mathbf{x}_0^\mathrm{K}$ at the last stage, and a consistent estimate of $\mathbf{x}_\mathrm{MMSE}$ can be computed by  averaging those samples, i.e. the empirical mean of samples converges in probability to $\mathbf{x}_\mathrm{MMSE}$ due to weak law of large numbers.
The variance of those samples is a solution to the error assessment for the reconstruction if
we trust the model parameterized by \cref{eq:learnt-tran} that is learned from a image database. The 95\% confidence
interval is computed for each pixel with its mean and variance.
Since a wider confidence interval (CI) means a larger margin of
error, the mean is overlaid with it to indicate the variability
of each pixel, and up to a certain point, the variability can
	cause a visual change on the image (cf. \ref{comp.fast}).


\section{Methods}
\subsection{Score Networks' Architecture}
The denoising score network is designed to predict the noise given an image degraded by Gaussian noise of a particular scale $\sigma_i$. To improve the quality of the predictions for different noise scales, we consider networks conditional on discrete and pseudo-continuous noise scales. The discrete one has a much larger gap between $\sigma_i$ and $\sigma_{i-1}$ than the pseudo-continuous one and usually has a smaller number of noise scales $N$, while the pseudo-continuous network is adaptive to a certain trained range of noise scales.
The sequence of noise scales 
$\{\sigma_i\}_{i=1}^\mathrm{N}$ is geometrically generated following the scheme in Ref. [\citen{Song_ICLR_2021}], i.e.
$\sigma_i =\sigma(\frac{i}{N})= \sigma_\mathrm{min}(\frac{\sigma_\mathrm{max}}{\sigma_\mathrm{min}})^\frac{i-1}{N-1}.$

For a discrete model, we add modified instance normalization layers that are conditional on the index of the noise scales following each convolution layer.
The conditional instance normalization \cite{Huang_ICCV_2017} is 
\begin{equation}
\hat{f}_{k}=\Phi[i, k] \frac{f_{k}-\mu_{k}}{s_{k}}+\Omega[i, k],
\end{equation}
where $\Phi \in \mathbb{R}^{N\times C}$ and $\Omega \in \mathbb{R}^{N\times C}$ are learnable parameters, $k$ denotes the index of a feature map $f_k$, $\mu_k$ and $s_k$ are the means and standard deviation over its spatial locations of the $k$-th feature map computed in each pass through the network, and $i$ denotes the index of $\sigma$ in $\{\sigma_i\}_{i=1}^\mathrm{N}$.

For a continuous model, we let networks be conditional on the index of noise scales by inserting random Fourier features  \cite{Rahimi_NIPS_2007}.
Three steps used to encode a noise index into random features are as follows:
\begin{itemize}
    \item Draw a random vector which has i.i.d. Gaussian $m$ entries with the specified standard deviation,
    \item Scale the random vector with the index $i$, then multiply it with $2\pi$,
    \item Apply sines and cosines to the scaled random vector, then concatenate them into $m\times 2$ matrix,
\end{itemize}
where $m$ is embedding size.
The encoded index is added to all the blocks listed in Supporting Table S1. 

With either one of the two modifications above, a network $\mathbf{s}_\theta(\mathbf{x}, i)$ has two inputs, i.e. noise corrupted image $\mathbf{x}$ and noise index $i$.
Real and imaginary parts of the images are interpreted as seperate channels when input into the neural network.
RefineNet \cite{Lin_CVPR_2017} is the backbone of all the score networks used in this work (cf. Supporting Figure S2).
Three variants from that are trained for different reconstruction experiments. The architectures of three networks are presented in detail in Supporting Table S1. We refer the readers to the codes available online for more information about them.
We labeled the three networks with $\mathtt{NET}_1, \mathtt{NET}_2,$ and $ \mathtt{NET}_3$, respectively, for ease of reference in the following.
$\mathtt{NET}_1$ is conditional on discrete noise scales, $\mathtt{NET}_2$ and $\mathtt{NET}_3$ are conditional on continuous noise scales.
We introduce self-attention modules into $\mathtt{NET}_3$ to capture long-range dependencies 
by adding non-local blocks as described previously\cite{Wang_Proc.Cvpr.IEEE._2018} so
that the network has the capability to model the dataset of high-resolution images.
\subsection{Dataset, Training and Inference}
We trained $\mathtt{NET}_1$ and $\mathtt{NET}_2$ on a dataset acquired by us already used and described in Ref. [\citen{Luo_Magn.Reson.Med._2020}]. $\mathtt{NET}_3$ was trained on a subset of the fastMRI dataset \cite{Zbontar_arXiv_2019}. Our dataset has
 1300 images containing T1-weighted, T2-weighted, T2-weighted {fluid-attenuated inversion recovery} (FLAIR), and T2$^*$-weighted brain images from 13 healthy volunteers examined with clinical standard-of-care protocols.
The brain images from fastMRI dataset \cite{Zbontar_arXiv_2019} were used for benchmark that contains T1-weighted (some with post contrast), T2-weighted and FLAIR images. For the detailed information of both dataset, we refer readers to corresponding publication.
Regarding the data partitioning, we first separated all multi-slice volumes into training and testing groups. 
Then we split the volume into two-dimensional slices (i.e., images).
Reference images - denoted $\mathbf{x}_0$ in the theory - were reconstructed from fully-sampled multi-channel k-space.
Then, these complex image datasets after coil combination were normalized to a maximum magnitude of 1. The coil sensitivity maps were computed with BART toolbox using ESPIRiT \cite{Uecker_Magn.Reson.Med._2014,uecker2020}. 
1300 images of size 256$\times$256 from the dataset used in Ref. [\citen{Luo_Magn.Reson.Med._2020}] were used to train $\mathtt{NET}_1$ and $\mathtt{NET}_2$. 1000 images were used for training, and 300 images were used for testing.
All networks are trained for 1000 epochs, i.e. iterations over all training images. For the training of $\mathtt{NET}_3$, we used the T2-weighted FLAIR contrast images of size 320$\times$320 that are reconstructed from fastMRI raw k-space data. 2937 images are for training, 326 images are for testing.

Three score networks are implemented with Tensorflow \cite{Abadi_OSDI_2016}. 
The hyperparameters used to train the three score networks are listed in Supporting Table S2.
With the trained networks, we implemented MCMC sampling Algorithm \ref{alg:seq} with Tensorflow and Numpy \cite{Harris_Nature_2020}, and then explored the posterior $p(\mathbf{x}|\mathbf{y})$ in different experimental settings.
We trained three score networks once separately for all the experiments we did in this work.
These three models can support all experiments performed in this study with variable undersampling patterns, coil sensitivity maps, channel numbers.
It took around 43 and 67 seconds, respectively, to train $\mathtt{NET}_1$ and $\mathtt{NET}_2$ for one epoch on one NVIDIA A100 GPU with 80GB. For $\mathtt{NET}_3$, it took around 500 seconds per epoch on two NVIDIA A100 GPUs using the multi-GPU support from Tensorflow.	In the spirit of reproducible research, codes and data to reproduce all experiments are made available\footnote{\url{https://github.com/mrirecon/spreco}}. 

\subsection{Experiments}
\textbf{Single Coil Unfolding:}  To investigate how the Markov chain explores the solution space of the inverse problem $\mathbf{y}=\mathcal{A}\mathbf{x} + \eta$, we designed the single coil unfolding experiment. The single channel k-space is simulated out of multi-channel k-space data. The odd lines in k-space are retained. 
10 samples were drawn from the posterior $p(\mathbf{x}|\mathbf{y})$. 
$\mathtt{NET}_1$ was used to construct transition kernels and the parameters in Algorithm \ref{alg:seq} are $\mathrm{K}=50,  \mathrm{N}=10, \lambda=6$. We redo the experiment with the object shifted to bottom.
This experiment has an inherent ambiguity which can not be resolved using the data alone and where the reconstruction is strongly determined by the prior. Thus, it mimics in a synthetic setting a situation with high undersampling where hallucinations were observed in the reconstruction of some deep-learning methods \cite{Muckley_IEEETrans.Med.Imag._2021}.

\bf{Multi-Coil Reconstruction}
Multi-channel data points from 
Cartesian k-space are randomly picked with variable-density poisson-disc
sampling and the central 20$\times$20 region is fully acquired. The acquisition mask covers 11.8\% k-space and the corresponding zero-filled reconstruction is shown \cref{fig:mul_chainb}.
We initialized 10 chains and the $\mathbf{x}_\mathrm{MMSE}$ was computed using different numbers of samples.
$\mathtt{NET}_1$ was used to construct transition kernels and the parameters in Algorithm \ref{alg:seq} are $\mathrm{K}=\delete{50}\b{30},  \mathrm{N}=\delete{10}\b{15}, \lambda=13$. \R{R.A1}
To visualize the process of sampling, we use peak-signal-noise-ratio (PSNR in dB) and similarity index (SSIM) as metrics to track intermediate samples.
The comparisons are made between the magnitude of $\mathbf{x}_\mathrm{MMSE}$ and the ground truth $\tilde{\mathbf{x}}$ after normalized with $\ell_2$-norm.

\bf{More Noise Scales}
To investigate how the number of noise scales influences the proposed method, we reconstructed the image from the undersampled k-space that was used in the multi-coil experiment.
$\mathtt{NET}_2$ was used to construct transition kernels and the parameters in Algorithm \ref{alg:seq} are $\mathrm{K}=5,  \mathrm{N}=70, \lambda=25 $.

\bf{Investigation of the Burn-in Phase}
To investigate the burn-in phase illustrated in \cref{fig:mcmc}, we
split up into multiple chains at a certain noise scale when drawing samples from the posterior $p(\mathbf{x}|\mathbf{y})$.
For instance, we denote by ($\mathbf{x}_\mathrm{MMSE}, 60$) the $\mathbf{x}_\mathrm{MMSE}$ that is computed with 10 samples drawn from $p(\mathbf{x}|\mathbf{y})$ by 
splitting up into
10 chains at the 60$^\mathrm{th}$ noise scale.
By changing the splitting
point, we got different sets of samples that are from chains of different length and computed the final $\mathbf{x}_\mathrm{MMSE}$ respectively.
We have two sets of $\mathbf{x}_\mathrm{MMSE}$ that are reconstructed from the undersampled k-space using two sampling patterns separately.
The central 20x20 region is obtained and the k-space, outside the center, is randomly picked up retrospectively (10\%, 20\%). 
$\mathtt{NET}_2$ was used to construct Markov transition kernels and the parameters in Algorithm \ref{alg:seq} are $\mathrm{K}=5,  \mathrm{N}=70, \lambda=25$. 

\bf{Investigation into MAP}
To verify the samples are located around the local modality of the posterior, we disabled the disturbance 
with noise after stochastic inference  with the last distribution $\tilde{p}(\mathbf{x}_{0} \mid \mathbf{x}_{1})$ and ran 200 iterations more to get extended samples. What's more, we repeated this procedure with determinate inference, in which the disturbance was disabled during sampling iterations to get one deterministic sample, i.e., MAP estimation.
 A Poisson-disc sampling pattern is generated without variable density and with 2-fold undersampling along phase and frequency encoding directions.
$\mathtt{NET}_2$ was used to construct transition kernels and the parameters in Algorithm \ref{alg:seq} are $\mathrm{K}=5,  \mathrm{N}=70, \lambda=25$. 

\bf{Comparison to $\ell_1$-regularized Reconstruction}
A comparison using the fastMRI dataset was used to evaluate the performance of the proposed method. We noticed that the raw k-space data is padded with zeros to make them have the same dimension. The effect caused by zero paddings is investigated in Ref. [\citen{Shimron_PNAS_2022}]. Since we only used the images that were reconstructed from the zero padded k-space for training, the issue caused by the synthesized k-space does not exist in our work. The undersampling pattern for each slice is randomly generated in all retrospective experiments.
$\mathtt{NET}_3$ was used to construct transition kernels.
The parameters in Algorithm \ref{alg:seq} are $\mathrm{K}=3,  \mathrm{N}=90, \lambda=20$ and 10 samples were drawn to compute $\mathbf{x}_\mathrm{MMSE}$. 
The data range for computing PSNR and SSIM is determined by the maximum over each slice.

\bf{Transferability}
To investigate the transferability of learned prior information from T2 FLAIR images to other contrasts, we acquired T1-weighted (TR=2000ms, TI=900ms, TE=9ms) and T2-weighted (TR=9000ms, TI=2500ms, TE=81ms) FLAIR k-space data using a 2D multi-slice turbo spin-echo sequence with a 16-channel head coil at 3T (Siemens, 3T Skyra).  $\mathtt{NET}_3$ (trained with T2 FLAIR images) was used to construct transition kernels. The parameters in Algorithm \ref{alg:seq} are $\mathrm{K}=5, \mathrm{N}=70, \lambda=20$.

\bf{Comparison to fastMRI challenge}
	\label{comp.fast}
	As a comparison to the unrolled neural network, the XPDNet \cite{Zaccharie_arXiv_2020}
	is selected as the reference which ranked 2nd in the fastMRI challenge.
	Two networks were trained for acceleration factors 4 and 8, using 
	retrospectively undersampled data from the fastMRI dataset [10] using
	equidistant Cartesian masks and the trained models that are publicly
	available\footnote{\url{https://huggingface.co/zaccharieramzi}}.
	For the proposed method, $\mathtt{NET}_3$ was used to construct transition
	kernels. The parameters in Algorithm \ref{alg:seq} are
	$\mathrm{K}=4,  \mathrm{N}=90, \lambda=20 $. The confidence interval after thresholding
	is used as the color map to indicate that a region has high uncertainty. Be consistent with the evaluation the XPDNet provided, 30 FLAIR volumes are used for validation to compute metrics.
\section{Results}
\subsection{Single Coil Unfolding}
\label{sec:single}
\begin{figure*}
	\labelphantom{fig:singlea}
	\labelphantom{fig:singleb}
	\labelphantom{fig:singlec}
	
		\centering
		\includegraphics[width=0.75\textwidth]{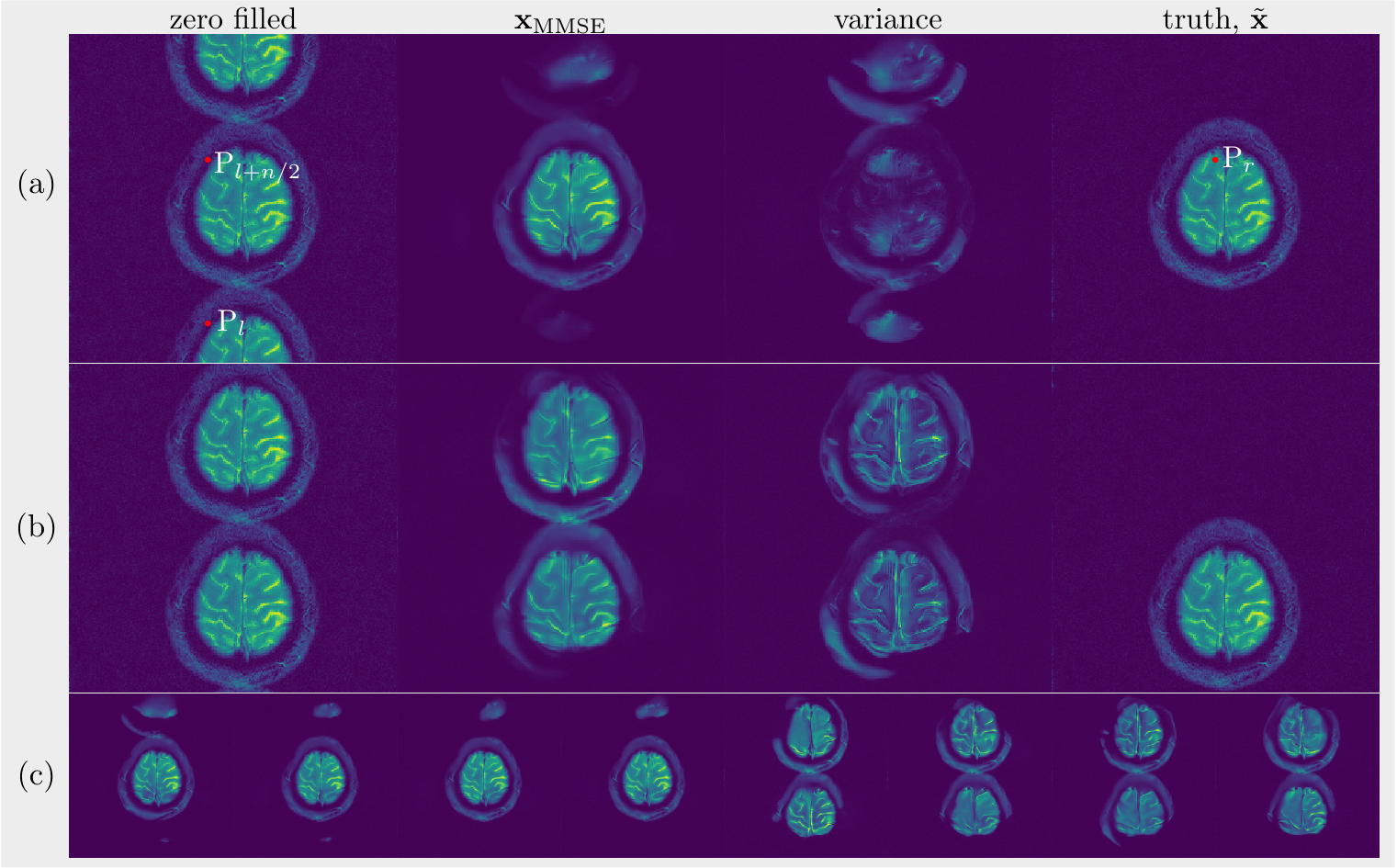}
		
	\caption{Single-coil unfolding with $\mathtt{NET}_1$. The k-space is undersampled by skipping every second line. Aliased images, $\mathbf{x}_\mathrm{MMSE}$, variance maps and ground truth are shown.
		(a) The object is centered. (b) The object is shifted.
	(c) Selected solutions are presented. The left four are centered and the right four are shifted.}
	\label{fig:single}
\end{figure*}

As expected, the lack of spatial information from coil sensitivities without parallel imaging leads to huge errors and folding artifacts still exist in $\mathbf{x}_\mathrm{MMSE}$ as shown in \cref{fig:single}. Since only odd lines are acquired, all images in which the superposition of points $\mathrm{P}_{l}$ and $\mathrm{P}_{l+2/n}$  equals to the points $\mathrm{P}_r$ in ground truth are solutions to $\mathbf{y}=\mathcal{A}\mathbf{x} +\epsilon$ with the same error (the residual norm $\|\mathbf{y}-\mathcal{A}\mathbf{x}\|^2$). Selected solutions are presented in \cref{fig:singlec}.
The variance map indicates the uncertainty of the solutions, which in this experiment is similar to the hallucinations observed in for some deep-learning methods for high undersampling\cite{Muckley_IEEETrans.Med.Imag._2021}.
 The errors of the estimation $\mathbf{x}_\mathrm{MMSE}$ are largely reduced compared to the zero-filled reconstruction because of prior knowledge from the learned reverse process (cf. \cref{fig:singlea}). The shift of the object increases the symmetry and then leads to even bigger errors as learned reverse process know less about images that were shifted (cf. \cref{fig:singleb}).

\subsection{Multi-Coil Reconstruction}
\label{sec:multi}
\begin{figure*}[t]
	\labelphantom{fig:mul_chaina}
	\labelphantom{fig:mul_chainb}
	\labelphantom{fig:mul_chainc}
	
	\centering
		\includegraphics[width=0.75\textwidth]{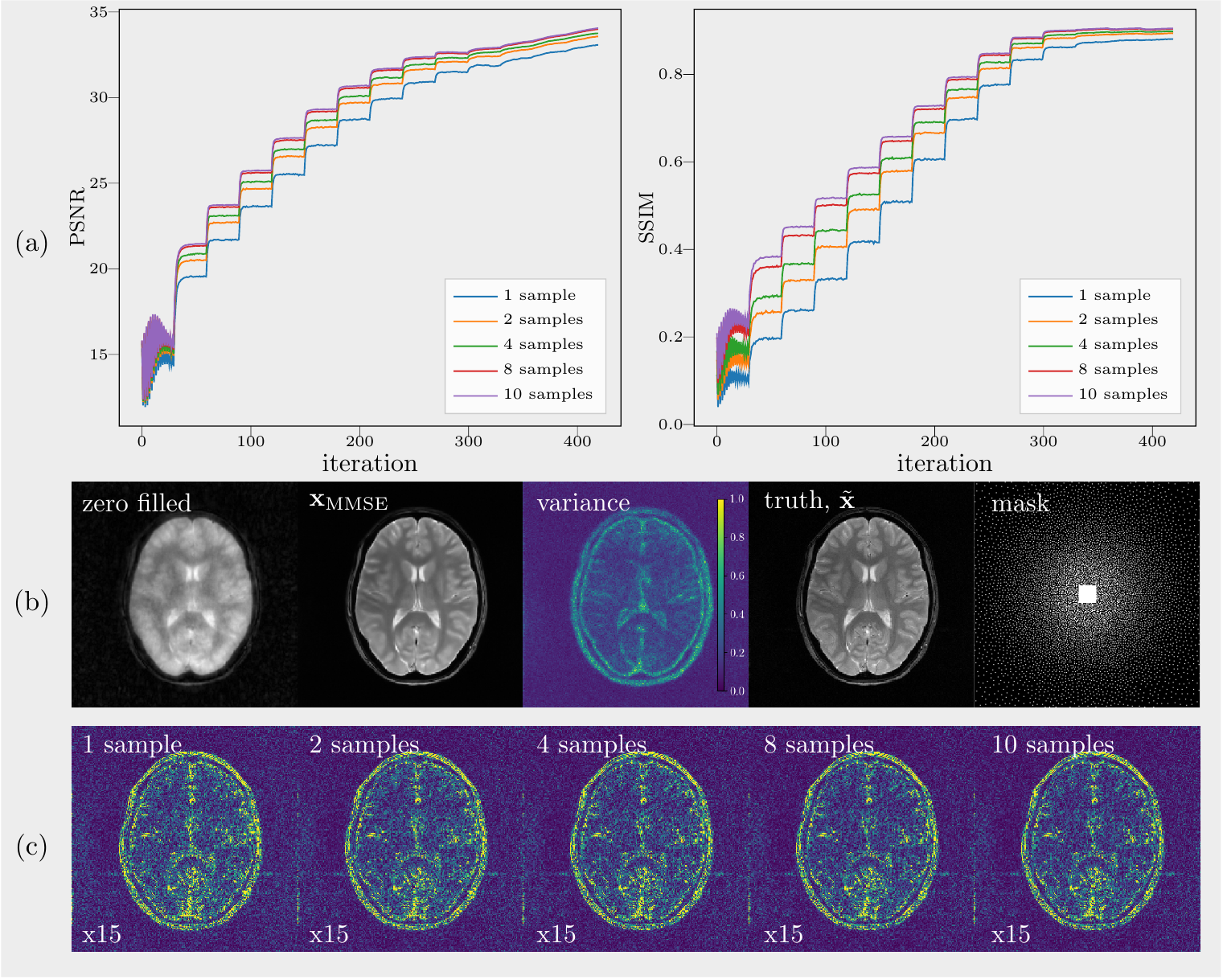}
		
	\caption{Multi-coil reconstruction with $\mathtt{NET}_1$. Results :
		(a) The curves of PSNR and SSIM over iterations 
		for $\mathbf{x}_\mathrm{MMSE}$s estimated by averaging a different number of samples
		(b) Zero-filled, $\mathbf{x}_\mathrm{MMSE}$, variance maps, truth and mask
		are presented. The final PSNR and the SSIM of $\mathbf{x}_\mathrm{MMSE}$ are
		\delete{36.06}\b{34.05}dB and \delete{0.9175}\b{0.9050}, respectively
		(c) The error maps between different $\mathbf{x}_\mathrm{MMSE}$s
		and the ground truth are presented.} \label{fig:mul_chain}
\end{figure*}

\cref{fig:mul_chain} shows the results for the multi-coil experiment. \cref{fig:mul_chaina} shows the evolution of the samples' PSNR and SSIM over the transitions of the data-driven Markov chain. Intermediate samples are presented in Supporting Figure S1. 
The convergence of samples at each noise level was reached as indicated by the PSNR and SSIM curves. 
When there are more samples, the $\mathbf{x}_\mathrm{MMSE}$ converges to higher PSNR and SSIM. In \cref{fig:mul_chainb}, 10 converged samples were used to compute $\mathbf{x}_\mathrm{MMSE}$ and the variance map.
Comparing with the ground truth, the variance map mainly reflects the edge information, which can be interpreted by the uncertainty that is introduced by the undersampling pattern used in k-space where many high frequency data points are missing but the low frequency data points are fully acquired.
In contrast to the single coil unfolding, the local spatial information from coil sensitivities reduces the uncertainties of missing k-space data.
Moreover, error maps qualitatively correspond to the variance map, with larger errors in higher variance regions as shown in \cref{fig:mul_chainc}. 
Lastly, the average over more samples leads to smaller error.
\subsection{More Noise Scales}
We also plotted the curve of PSNRs and SSIMs over iterations in \cref{fig:more-levelsa}
for $\mathtt{NET}_2$ which uses continuous noise scales.
The PSNR and SSIM of $\mathbf{x}_\mathrm{MMSE}$, which is computed with 10 samples, are 37.21dB and 0.9360, respectively.
Two $\mathbf{x}_\mathrm{MMSE}$ reconstructed separately with the application of $\mathtt{NET}_1$ and $\mathtt{NET}_2$ are presented in \cref{fig:more-levelsb} and variance maps are presented as well. 
The variance of the samples that are drawn with $\mathtt{NET}_2$ is less than those drawn with $\mathtt{NET}_1$, which means that we are more confident about the reconstuction using $\mathtt{NET}_2$. 
When we zoom into the region that has more complicated structures, the boundaries between white matter and gray matter are more distinct in the image recovered with $\mathtt{NET}_2$ and the details are more obvious, as shown in \cref{fig:more-levelsc}.
Hence, increasing the number of noise scales 
in $\mathtt{NET}_2$ relative to $\mathtt{NET}_1$
reduces the number of iterations and improves the quality of reconstruction using score networks of comparable size.
More noise scales make chains constructed with $\mathtt{NET}_2$ exploit the prior knowledge from training image dataset more effectively than chains constructed with $\mathtt{NET}_1$ which has fewer noise scales.
\begin{figure*}
	\centering
	\labelphantom{fig:more-levelsa}
	\labelphantom{fig:more-levelsb}
	\labelphantom{fig:more-levelsc}
		\includegraphics[width=0.87\textwidth]{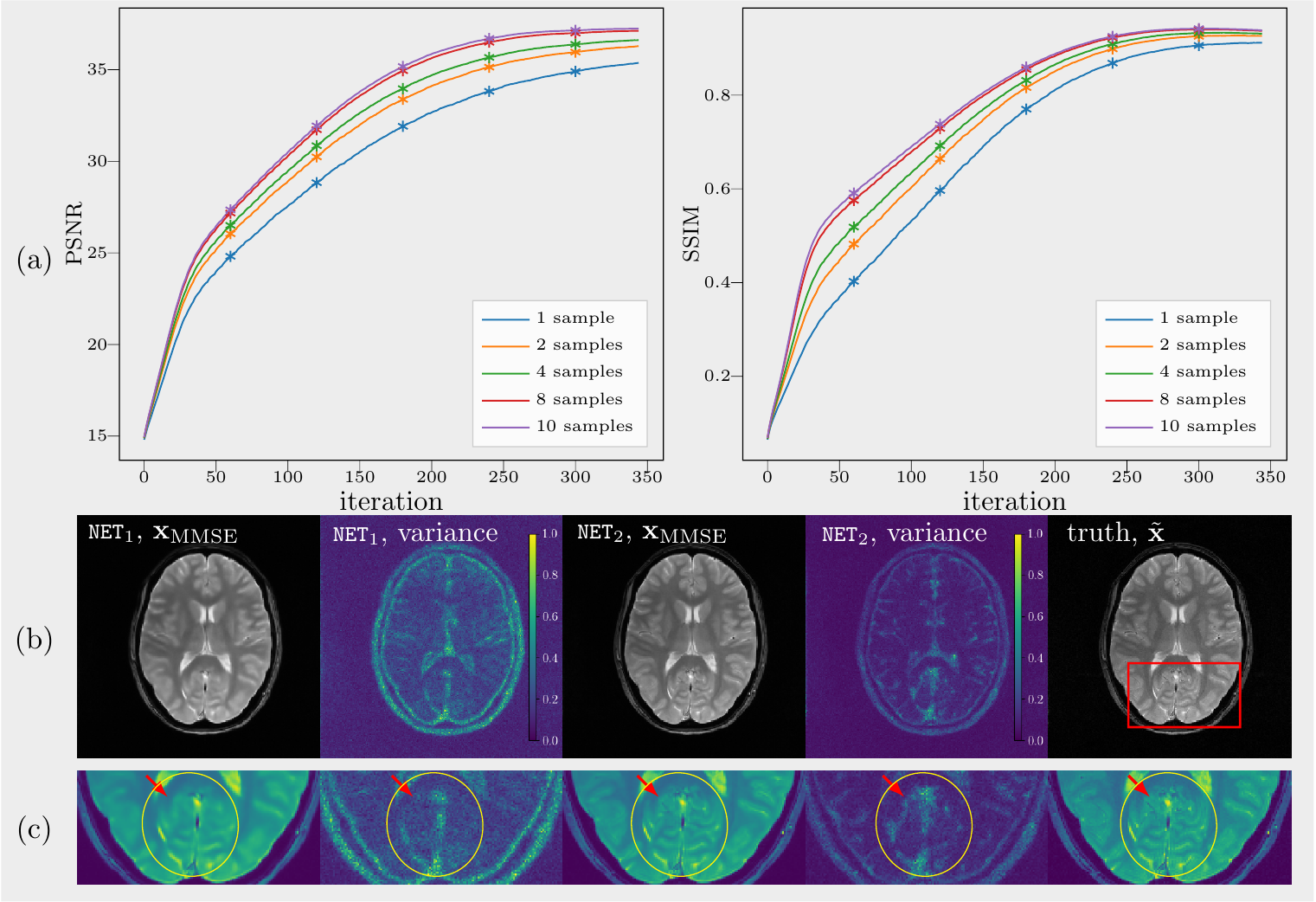}
	\caption{Effect of using continuous noise scales in $\mathtt{NET_2}$.
	(a) The convergence curves of PSNR and SSIM over iterations for $\mathtt{NET}_2$.
	(b) 	Reconstructed MMSE and variance maps for $\mathtt{NET}_2$ and $\mathtt{NET}_2$.
	(c) 	Zoomed view of selected structures (yellow circle, red arrow).}
	\label{fig:more-levels}
\end{figure*}
\subsection{Investigation of the Burn-in Phase}
\label{sec:4.4}
\begin{figure*}
\labelphantom{fig:burn-ina}
\labelphantom{fig:burn-inb}
\centering
		\includegraphics[width=0.87\textwidth]{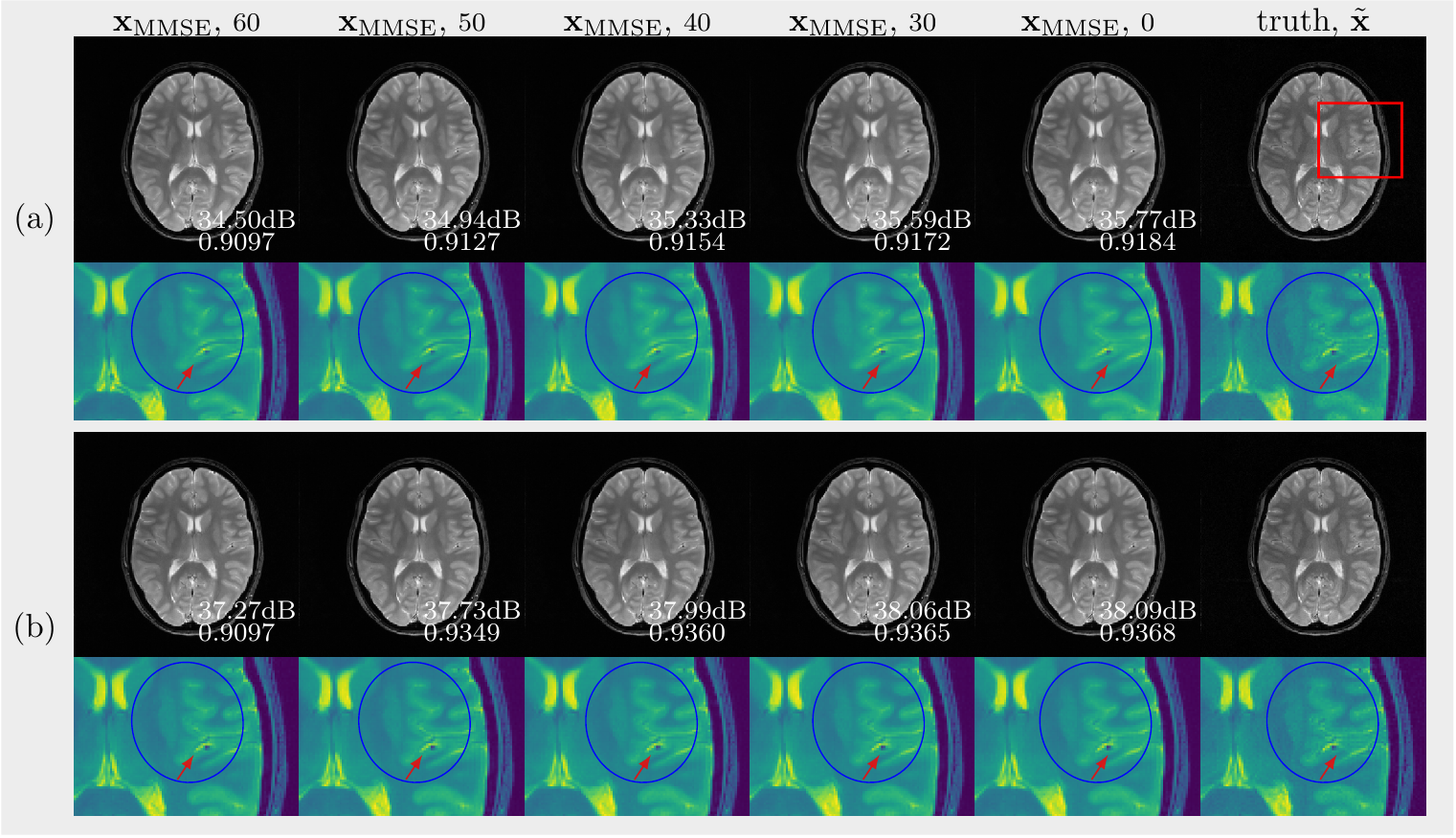}
	\caption{To investigate the burn-in phase the effect of splitting
	chains at different time points is shown for $\mathtt{NET}_2$ for reconstruction with
	(a) 10\% k-space data points and (b) 20\% k-space data points.}
\label{fig:burn-in}
\end{figure*}

The two sets of $\mathbf{x}_\mathrm{MMSE}$ are presented in \cref{fig:burn-in}. In \cref{fig:burn-ina}, the earlier we split chains, the closer the $\mathbf{x}_\mathrm{MMSE}$ gets to the truth. Especially, when we zoom into the region that has complicated structures (indicated by the red rectangle), the longer chains make fewer mistakes. The slightly distorted structure is seen in ($\mathbf{x}_\mathrm{MMSE}, 60$) highlighted with blue circles. The distortion has disappeared in ($\mathbf{x}_\mathrm{MMSE}, 0$) but some details are still missing. However, given more k-space data points, the longer chains do not cause a huge visual difference in the $\mathbf{x}_\mathrm{MMSE}$ as shown in \cref{fig:burn-inb}, even though there is a slight increase in PSNR and SSIM.
Although fewer data points mean more uncertainties, longer chains permit better exploration of the solution space, as shown by this experiment. Here, the image ($\mathbf{x}_\mathrm{MMSE}$, 60) took about one fourth of the time (4 minutes and 30 seconds) to compute than the image ($\mathbf{x}_\mathrm{MMSE}$, 0). For moderate undersampling rates, a burn-in phase is recommended for reducing computation time. 

\subsection{Investigation of the MAP}
\begin{figure*}
	\labelphantom{fig:mapa}
	\labelphantom{fig:mapb}
	\labelphantom{fig:mapc}
	\centering
		\includegraphics[width=0.8\textwidth]{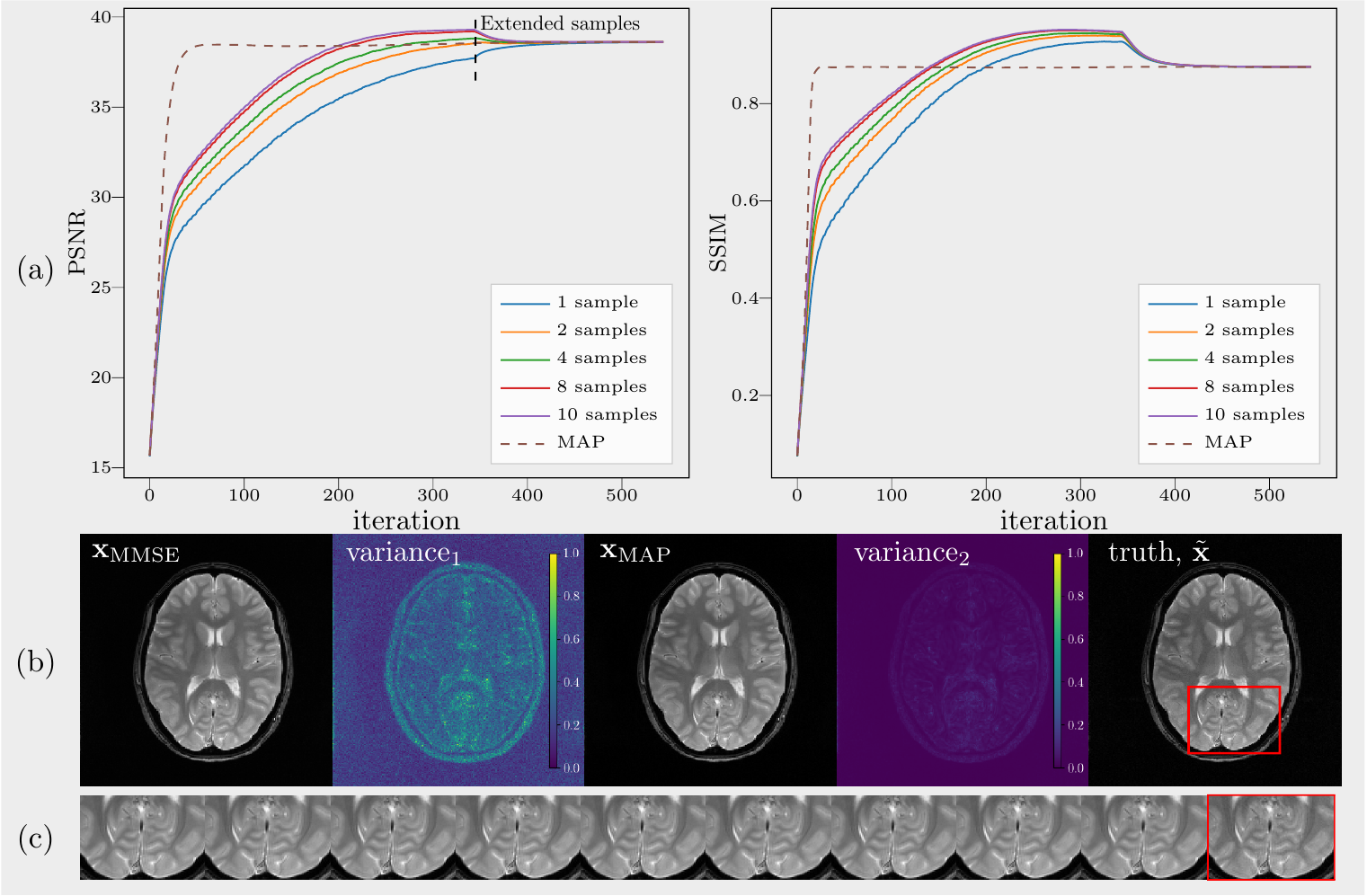}
	\caption{Investigation of the MAP reconstruced with $\mathtt{NET}_2$. 200 extended iterations after random exploration versa a deterministic estimate of MAP that are indicated by solid and dashed lines respectively. (a) The curves of PSNR and SSIM over iterations. (b) The sub-figure variance$_1$ and variance$_2$ were computed from unextended samples and extended samples respectively. $\mathbf{x}_\mathrm{MAP}$ is an extended sample. (c) The zoom-in region of 9 extended samples and the ground truth.}
	\label{fig:map}
\end{figure*}

In \cref{fig:map}, we plotted the curves of PSNR and SSIM over extended iterations for $\mathtt{NET}_2$
and presented reconstructions that are from the MMSE and MAP estimator.
As indicated by zoom-in images  and curves in Figure 7a and 7c,
 the extended samples converge to a consistent estimate of the MAP.
Measured by PSNR and SSIM, the MAP has better quality than individual samples.
As expected,  the MMSE obtained from averaging ten (non-extended) samples has better PSNR and SSIM than the MAP.
\subsection{Comparison to $\ell_1$-regularized Reconstruction} 
\begin{figure}
	\centering
		\includegraphics[width=\columnwidth]{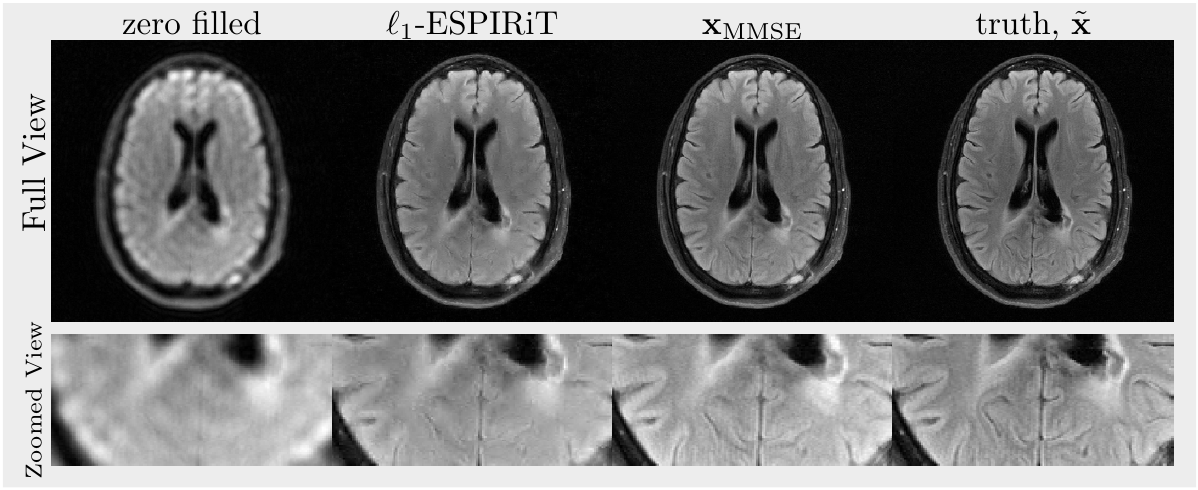}
	\caption{Comparison of the MMSE computed with $\mathtt{NET}_3$ to the $\ell_1$-wavelet regularized and zero-filled reconstruction. The high resolution image (320$\times$320) was reconstructed from k-space data using 10-fold undersampling. The regularization parameter was set to 0.01.}
	\label{fig:ben}
\end{figure}

The reconstructions with different methods are presented in \cref{fig:ben}. 
$\ell_1$-ESPIRiT denotes the reconstruction with the $\mathtt{pics}$ command of BART toolbox using $\ell_1$-wavelet regularization (0.01), which mostly recovers general structures while smoothing out some details. In $\mathbf{x}_\mathrm{MMSE}$, the majority of details are recovered, and the texture is almost identical to the ground truth, although some microscopic structures are still missing.
Each subject has 16 slices and the metrics of 3 subjects presented in Supporting Table S3 are the average over slices of each subject.
It's worth mentioning that PSNR and SSIM are influenced by the value-range of a slice in the evaluation of MR images.

\subsection{Transferability}
\begin{figure}

		\includegraphics[width=\columnwidth]{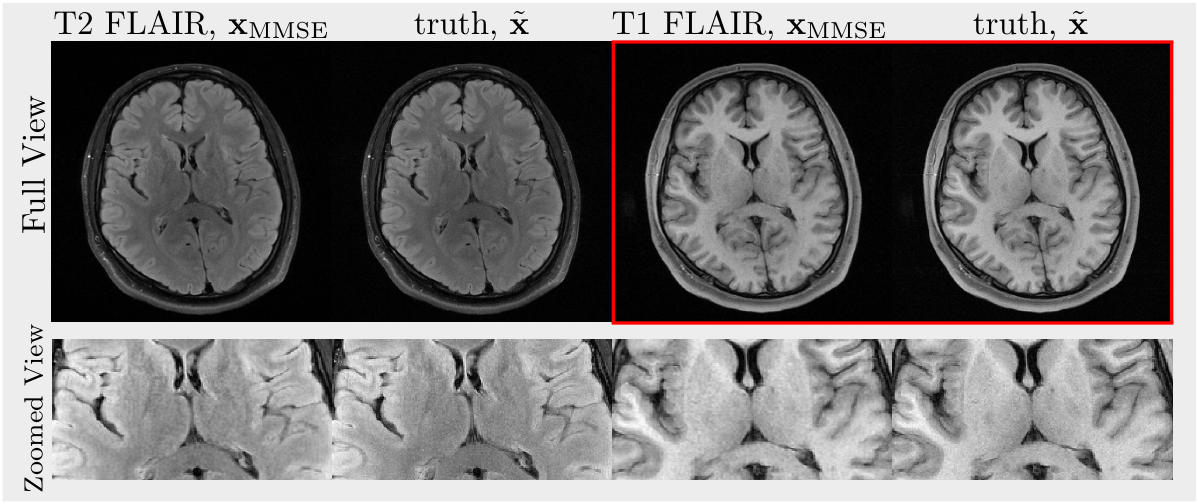}

	\caption{Transferability:
	Reconstruction of T2 and T1 FLAIR images (red box) using a Poisson-disc pattern with 8x undersampling in k-space
		using $\mathtt{NET}_3$ trained on T2 FLAIR images.}
	\label{fig:transfer}
\end{figure}

\cref{fig:transfer} shows a $\mathtt{NET}_3$ trained with T2 FLAIR contrast used to reconstruct a T1 FLAIR image (red box) in comparison to a T2 FLAIR image. No loss of quality can be observed.

\subsection{Comparison to fastMRI challenge}
	
	As discussed in Ref. [\citen{Arvinte__2021}], the ground
	truth matters when computing comparison metrics.
	We plotted the metrics of 30 volumes against a root sum of squares (RSS) and a
	coil combined image (CoilComb)
	in Supporting Figure S3, which shows XPDNet favors RSS that was used as labels for training it while
	$\mathbf{x}_\mathrm{MMSE}$ favors the other. Besides, the
	data range can be determined slice by slice or volume by volume, and the influences of that are not ignorable.

\begin{figure*}
	\centering
	\includegraphics[width=0.9\textwidth]{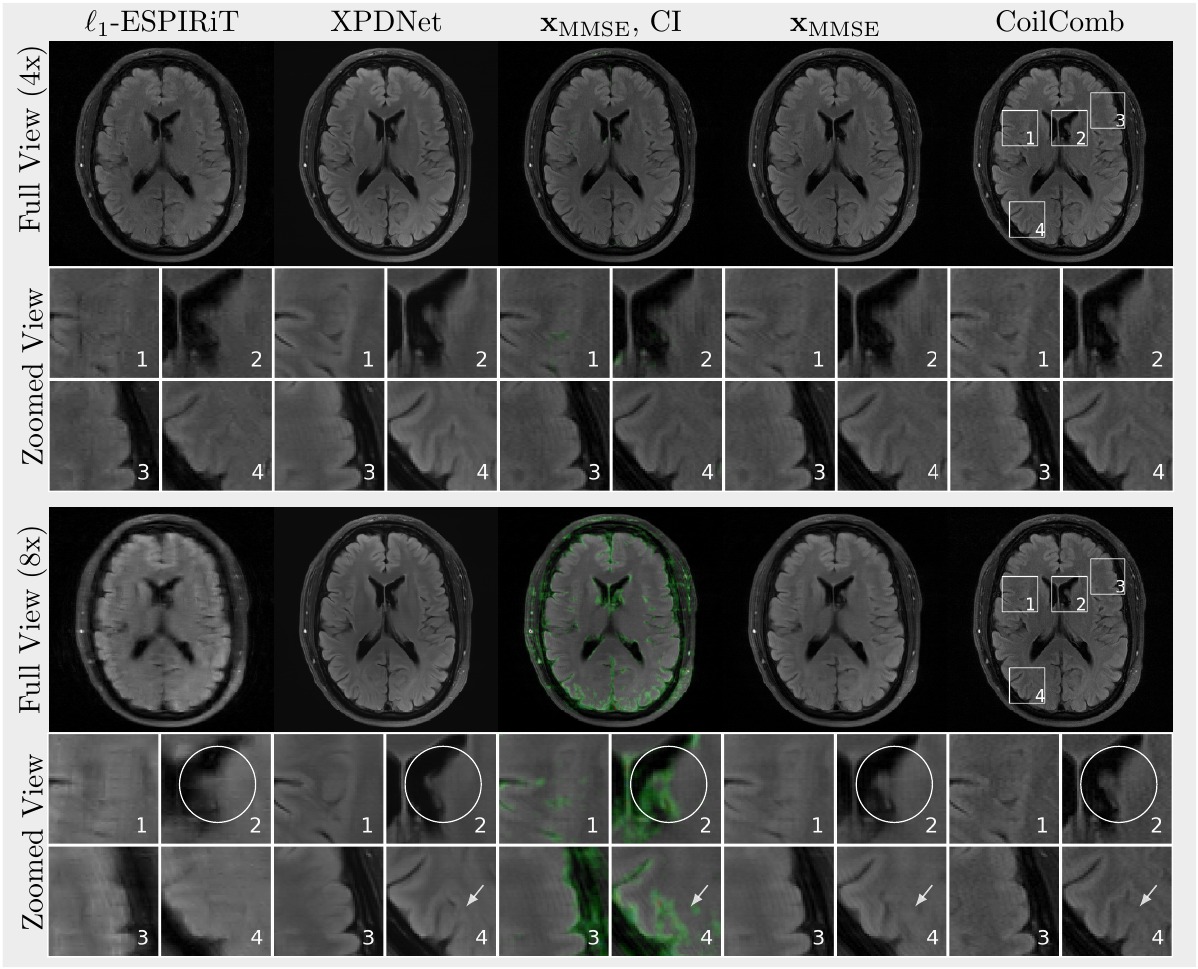}
		\caption{Comparison to fastMRI challenge. From the leftmost to rightmost column, reconstructions are $\ell_1$-ESPIRiT, XPDNet, $\mathbf{x}_\mathrm{MMSE}$ highlighted with confidence interval, $\mathbf{x}_\mathrm{MMSE}$ and a fully-sampled coil-combined image (CoilComb). Hallucinations appear when using 8-fold acceleration along the phase-encoding direction (horizontal) and are highlighted with the confidence interval after thresholding. Selected regions of interests are presented in a zoomed view.}
		\label{fig:comp_fast}
\end{figure*}

	Both methods provide nearly aliasing-free
	reconstruction at 4 or 8-fold acceleration. However,
	the hallucinations appear when using 8-fold 
	acceleration, highlighted with the green color (cf. \cref{fig:comp_fast}).

	All in all, a deep learning-based
	method has enough capability to generate a realistic-looking
	image even when the problem is highly underdetermined 
	as a result of undersampling, but the uncertainties
	inside it cannot be ignored.

\section{Discussion}
Generally, the Bayesian statistical approach provides a foundation for sampling the posterior $p(\mathbf{x}|\mathbf{y})$ and a natural mechanism for incorporating the prior knowledge that is learned from images. The generative model is used to construct Markov chains to sample the posterior.
The utilization of probabilistic generative models allows: 1) flexibility for changing the forward model of measurement; 2) exact sampling from the posterior term $p(\mathbf{x}|\mathbf{y})$; and 3) the estimation of uncertainty due to limited k-space data points.

\bf{Uncertainties of Reconstruction}
One advantage of the proposed approach over classical deterministic regularization methods is that it allows the quantification of uncertainties of the reconstruction with the variance map. That requires MCMC sampling technique.
The loss of spatial information of coils leads to the failure of unfolding, as demonstrated in Section \ref{sec:single}.
	High undersampling implies a high uncertainty about the solution, which may lead to hallucinations as observed in Ref. [\citen{Muckley_IEEETrans.Med.Imag._2021}] and \cref{fig:comp_fast}.
The regions with aliasing correspond to the high variance areas of the uncertainty map. 
With multiple coils, the reduction of high frequency data points in k-space leads to the loss of fine details, as demonstrated in Section \ref{sec:multi}. The $\mathbf{x}_\mathrm{MMSE}$ represents the reconstruction with minimum mean square error and the variance map evaluates the confidence interval of $\mathbf{x}_\mathrm{MMSE}$. \b{Furthermore, it is possible to derive error bounds from the variance of the posterior as reported Ref. [\citen{Narnhofer_arXiv_2022}].}

\bf{Overfitting and Distortion}
The proposed algorithm is an iterative refining procedure that starts from generating coarse samples with rich variations under large noise, before converging to fine samples with less variations under small noise.
For early iterations of the algorithm, each parameter update mimics stochastic gradient descent; however, as the algorithm approaches a local minimum, the gradient shrinks  and the chain produces the samples from the posterior.
	Lastly, we noticed that the balance between the learned transition and the data consistency plays an important role generally in the generation of realistic samples; here we refer readers to Supporting Figure S4. The larger $\lambda$, the stronger the consistency of data. \b{Besides, we found that a large value of $\text{K}$ is required for using the discrete noise conditional score
network in \cref{alg:seq} while a smaller value is sufficient for the continuous noise scales. While the $\text{N}$ in \cref{alg:seq} is larger for the continuous case, the total number of iterations in both cases is comparable. }\R{R3.2}

\bf{Computational Burden}
The promising performance of this method comes at the price of demanding computation. It takes around 10 minutes to reproduce the results in \cref{fig:more-levels} while $\ell_1$-ESPIRiT takes about 5 seconds with BART for a single slice. The possible solutions to the computation burden are to: 1) accelerate the inferencing of neural networks; 2) parallelize the sampling process when multiple chains are used; and 3) reduce the number of iterations using more efficient MCMC sampling techniques. Furthermore, reducing the scale of networks is also viable. The introduction of burn-in experiment in \cref{sec:4.4} is a direct way to overcome this shortcoming when the undersampling factor is moderate.

\bf{Relationship to Generative Models}
To our knowledge, the construction of image models to exploit prior knowledge was first introduced in Ref. [\citen{Geman_IEEETrans.PatternAnal.MachineIntell._1984}] in which the handcrafted model which extracts edge information was used for image restoration. Following that framework, the learned generic image priors from generative perspective are investigated in Ref. [\citen{Song_ICLR_2021, Roth_Proc.Cvpr.IEEE._2005, Schmidt_Proc.Cvpr.IEEE._2010}], which permits more expressive modeling. In the medical imaging field, image priors learned  with variational autoencoder \cite{Tezcan_IEEETrans.Med.Imag._2018, Kingma_ICLR_2014} and PixelCNN \cite{Luo_Magn.Reson.Med._2020, Salimans_ICLR_2017} were applied to MRI image reconstruction. As a comparison to the method in Ref. [\citen{Luo_Magn.Reson.Med._2020}], the result is presented in Supporting Figure S5.
Compared with some unrolled network based deep learning image reconstruction methods, the application of image priors is independent of k-space data and coil sensitivities, which permits a more versatile use of the method using different k-space acquisition strategies.

\bf{Limitations}
PSNR and SSIM only give a partial and distorted view of image quality. The influence
of the ground truth and noise properties of the background have a severe influence, as
does the selected data range used for computing the metrics. Thus, rating of image quality
by human readers would be an important next step in the evaluation of the technique.
Also the clinical usefulness of the uncertainty maps requires further investigations.
\b{To facilitate the use in clinical studies, we implemented the sampling in
the BART toolbox.\cite{Luo__2021a,Blumenthal_Magn.Reson.Med._2023}}

\section{Conclusion}

The proposed reconstruction method combines concepts from machine learning, Bayesian inference and image reconstruction.
In the setting of Bayesian inference, the image reconstruction is realized by drawing samples from the posterior term $p(\mathbf{x}|\mathbf{y})$ using data-driven Markov chains,
providing a minimum mean square reconstruction and uncertainty estimation. The prior information can be learned from an existing image database,where the generic generative priors based on the diffusion process allow for flexibility regarding contrast, coil sensitivities, and sampling pattern.

\section*{Acknowledgement}
We acknowledge funding by the "Niedersächsisches Vorab" funding line of the Volkswagen Foundation.
We would like to thank Xiaoqing Wang for his help in preparing this manuscript as well as Christian Holme for 
help with our computer systems.

\bibliography{radiology}
\vfill\pagebreak
\section*{Supporting Information}
The following supporting information is available as part of the online article:
\vskip\baselineskip\noindent
\noindent
\textbf{Figure S1.} {Samples and $\mathbf{x}_\mathrm{MMSE}$ from intermediate distributions.}

\noindent
\textbf{Figure S2.} {Overview of RefineNet and refine blocks.}

\noindent
\textbf{Figure S3.} {PSNR and SSIM Metrics for different ground truths and data ranges.}

\noindent
\textbf{Figure S4.} {The impact on the reconstruction when changing $\lambda$.}

\noindent
\textbf{Figure S5.} {The comparison to the deep Bayesian reconstruction method in Ref. [\citen{Luo_Magn.Reson.Med._2020}].}

\noindent
\textbf{Table S1.} {The architectures of networks used in this work.}

\noindent
\textbf{Table S2.} {The hyperparameters used for training networks.}

\noindent
\textbf{Table S3.} {The metrics from the comparison between $\ell_1$-ESPIRiT and $\mathbf{x}_\mathrm{MMSE}$.}

\appendix
\section{Rewrite in terms of posterior}
\label{appendix.a}
Because the forward diffusion is a Markov process and start at $\mathbf{x}_{0}$, with Bayes' rule we have
\begin{equation}
    q\left(\mathbf{x}_{i} \mid \mathbf{x}_{i-1}, \mathbf{x}_{0}\right) = q\left(\mathbf{x}_{i-1} \mid \mathbf{x}_{i}\right) \frac{q\left(\mathbf{x}_{i} \mid \mathbf{x}_{0}\right)}{q\left(\mathbf{x}_{i-1} \mid \mathbf{x}_{0}\right)}.
    \label{eq:ba}
\end{equation}
Substituting density function into \cref{eq:ba} yields
\begin{align}
    q(\mathbf{x}_{i-1}|\mathbf{x}_i, \mathbf{x}_0) & = q(\mathbf{x}_i|\mathbf{x}_{i-1})\cdot\frac{ q(\mathbf{x}_{i-1}| \mathbf{x}_0)}{q(\mathbf{x}_i| \mathbf{x}_0)}\\
    &=\frac{1}{\sqrt{(2\pi\beta_i^2)^{N_p}}}\cdot \frac{b_i^{2N_p}}{b^{2N_p}_{i-1}}\exp \Bigl[-\Bigl(\frac{\|\mathbf{x}_i-\mathbf{x}_{i-1}\|^2}{\beta_i^2} \nonumber \\
    & +\frac{\|\mathbf{x}_{i-1}-\mathbf{x}_0\|^2}{b_{i-1}^2}-\frac{\|\mathbf{x}_{i}-\mathbf{x}_0\|^2}{b_{i}^2}\Bigr)\Bigr].
\end{align}

Let $\frac{b_{i-1}^2+\beta_i^2}{b_i^2}=1$, which is satisfied with \cref{eq:5}, we have
\begin{align}
        q(\mathbf{x}_{i-1}|\mathbf{x}_i, \mathbf{x}_0) & =\frac{1}{\sqrt{(2\pi\beta_i^2)^{N_p}}}\cdot \frac{b_i^{2N_p}}{b^{2N_p}_{i-1}} \exp \Bigl[-\Bigl(\frac{\|\mathbf{x}_i-\mathbf{x}_{i-1}\|^2}{\beta_i^2}\nonumber\\
        &+\frac{\|\mathbf{x}_{i-1}-\mathbf{x}_0\|^2}{b_{i-1}^2}-\frac{\|\mathbf{x}_{i}-\mathbf{x}_0\|^2}{b_{i}^2}\Bigr)\Bigr]\\
         &=\frac{1}{\sqrt{(2\pi\beta_i^2)^{N_p}}}\cdot \frac{b_i^{2N_p}}{b^{2N_p}_{i-1}}\exp \Bigl[-\Bigl(\frac{\|\mathbf{x}_{i-1}-\boldsymbol{\mu}\|^2}{\beta_i^2\cdot\frac{b_{i-1}^2}{b_i^2}}\Bigr)\Bigr],
    \end{align}
where
\begin{equation}
    \boldsymbol{\mu}=\frac{b_{i-1}^2}{b_i^2}\cdot \mathbf{x}_i+\frac{\beta_i^2}{b_i^2}\cdot \mathbf{x}_0.
\end{equation}
\section{KL divergence of two Gaussian distributions}
\label{appendix.f}
Let $p(\mathbf{x}) = \mathcal{CN}(\boldsymbol{\mu}_1, \sigma_1^2\mathbf{I})$ and $q(\mathbf{x}) = \mathcal{CN}(\boldsymbol{\mu}_2, \sigma_2^2\mathbf{I})$ and the KL divergence is defined by
\begin{equation}
    \displaystyle D_{\text{KL}}(P\parallel Q)=\int _{-\infty }^{\infty }p(\mathbf{x})\log \bigl({\frac {p(\mathbf{x})}{q(\mathbf{x})}}\bigr)\,d\mathbf{x}. \nonumber
\end{equation}
Therefore,
\begin{align}
    D_{\text{KL}}(P\parallel Q)&= \int [\log p(\mathbf{x}) - \log q(\mathbf{x})] p(\mathbf{x}) d\mathbf{x} \nonumber\\
    &=\mathbb{E}_{p(\mathbf{x})} \bigl[N_p \log\bigl(\frac{\sigma_2}{\sigma_1}\bigr) + \frac{1}{\sigma_2^2}  \|\mathbf{x}-\boldsymbol{\mu}_2\|^2 - \frac{1}{\sigma_1^2} \|\mathbf{x}-\boldsymbol{\mu}_1\|^2 \bigr]\nonumber\\
    &=N_p\cdot \log\bigl(\frac{\sigma_2}{\sigma_1}\bigr) + \frac{1}{\sigma_2^2} \mathbb{E}_{p(\mathbf{x})} \bigl[\|\mathbf{x}-\boldsymbol{\mu}_2\|^2\bigr] - {1}. \nonumber
\end{align}
where $N_p$ is the dimensionality $n\times n \times 2$. Noting that
\begin{equation*}
	\|\mathbf{x} - \boldsymbol{\mu}_2\|^2 = 
	\|\mathbf{x}-\boldsymbol{\mu}_1\|^2 + 2\mathbb{Re}(\mathbf{x}-\boldsymbol{\mu}_1)^H(\boldsymbol{\mu}_1-\boldsymbol{\mu}_2) + \|\boldsymbol{\mu}_1-\boldsymbol{\mu}_2\|^2
\end{equation*}
we arrive at
\begin{align*}
	D_{\text{KL}}(P\parallel Q) 
	=& N_p\cdot\log\bigl(\frac{\sigma_2}{\sigma_1}\bigr) + \frac{1}{\sigma_2^2}
		\bigl(\mathbb{E}_{p(\mathbf{x})}\bigl[\|\mathbf{x}-\boldsymbol{\mu}_1\|^2\bigr] \\
	 & + 2\mathbb{Re}(\boldsymbol{\mu}_1-\boldsymbol{\mu}_2)^H\mathbb{E}_{p(\mathbf{x})}\bigl[\mathbf{x}-\boldsymbol{\mu}_1\bigr] \\
	 & + \mathbb{E}_{p(\mathbf{x})}\bigl[|\boldsymbol{\mu}_1-\boldsymbol{\mu}_2\|^2\bigl]\bigr) - {1} \\
	=& N_p\cdot \log\bigl(\frac{\sigma_2}{\sigma_1}\bigr) + \frac{\sigma_1^2 + \|\boldsymbol{\mu}_1-\boldsymbol{\mu}_2\|^2}{\sigma_2^2} - {1}.
\end{align*}

	\vfill\pagebreak\phantom{text}\vfill\pagebreak
	

\section*{Supplementary material}

## SUPPORTING INFORMATION

## Grid of Samples

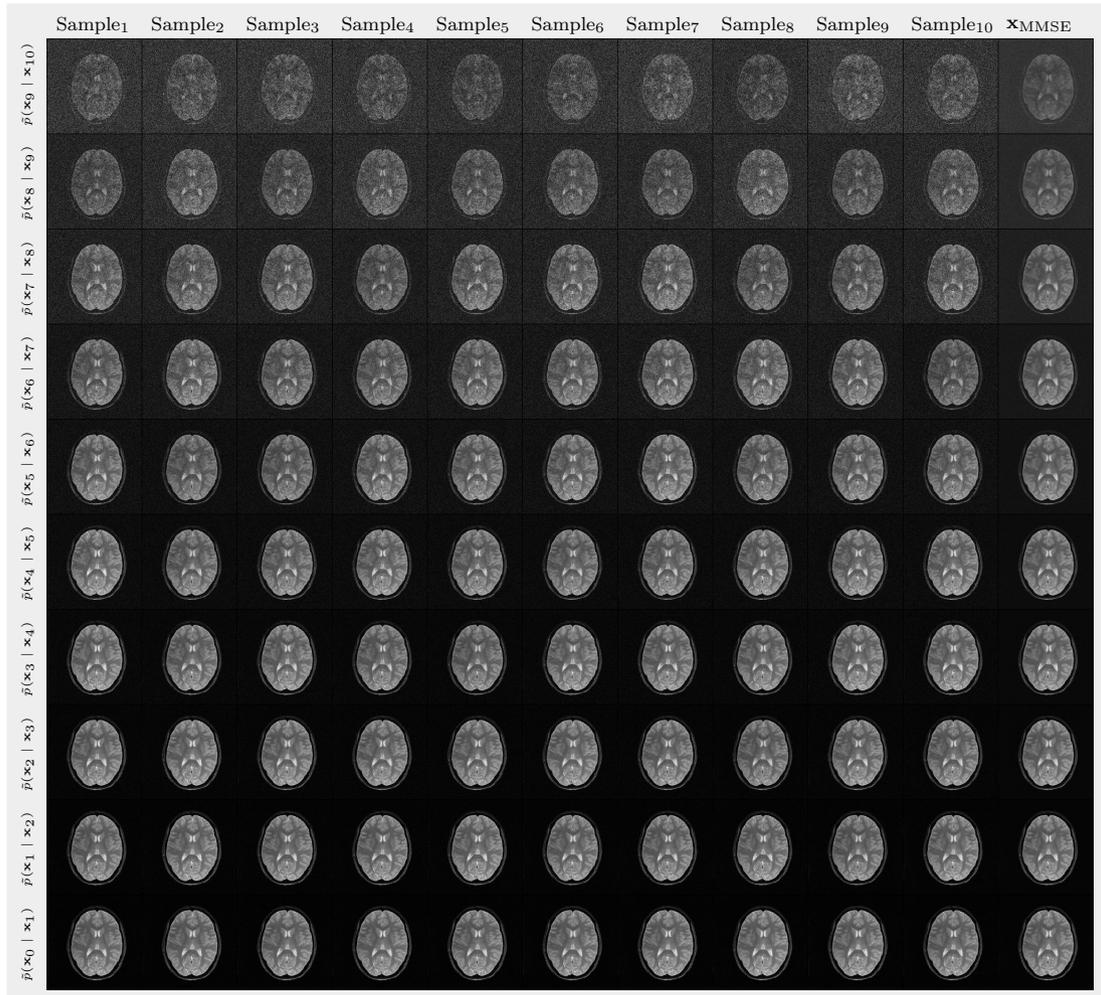

**FIGURE S1** : Samples and $\mathbf{x}_{\mathrm{MMSE}}$ from intermediate distributions are presented here. Each $\mathbf{x}_{\mathrm{MMSE}}$ is the average over 10 samples.

## Implementation Details of Networks

The architecture of RefineNet[36] is U-Net based as shown below. The *ref*, *res*, *rcu*, *msf* and *crp* are abbreviations of refine block, residual block, residual convolution unit, multi-resolution fusion and chained residual pooling. All the blocks used in the neural networks are conditional on noise scales. For a discrete score network, the conditioning is achieved by instance normalization layers, while in a continuous score network, the conditioning is achieved by adding sinusoidal position embeddings. Details are listed in below Table S1.

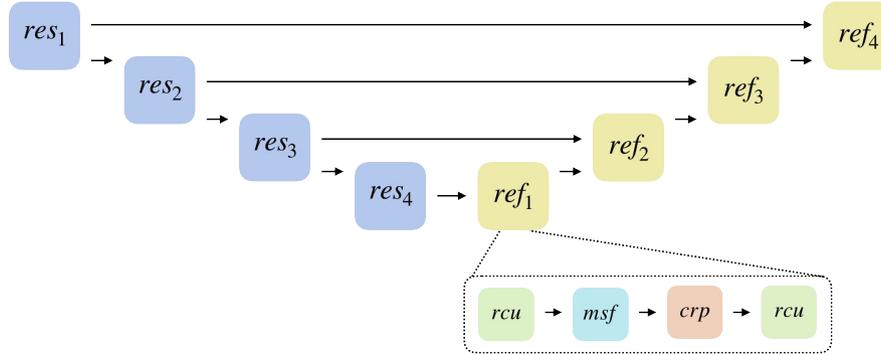

**FIGURE S2** : Overview of RefineNet and refine blocks.

## PSNR and SSIM Metrics for Different Ground Truths and Data Ranges

PSNR is defined via the mean squared error (MSE) and the maximum possible pixel value of the image (MAX)

$$\text{PSNR} = 20 \cdot \log_{10}(\text{MAX}) - 10 \cdot \log_{10}(\text{MSE}).$$

As shown in the following figure, the PSNR metric is influenced by the ground truth, i.e., RSS or CoilComb, and whether the maximum is computed over the slice or volume. The RSS of the volumes are the labels used to train XPDNet and the samples used to compute $\mathbf{x}_{\text{MMSE}}$ are generated slice by slice.

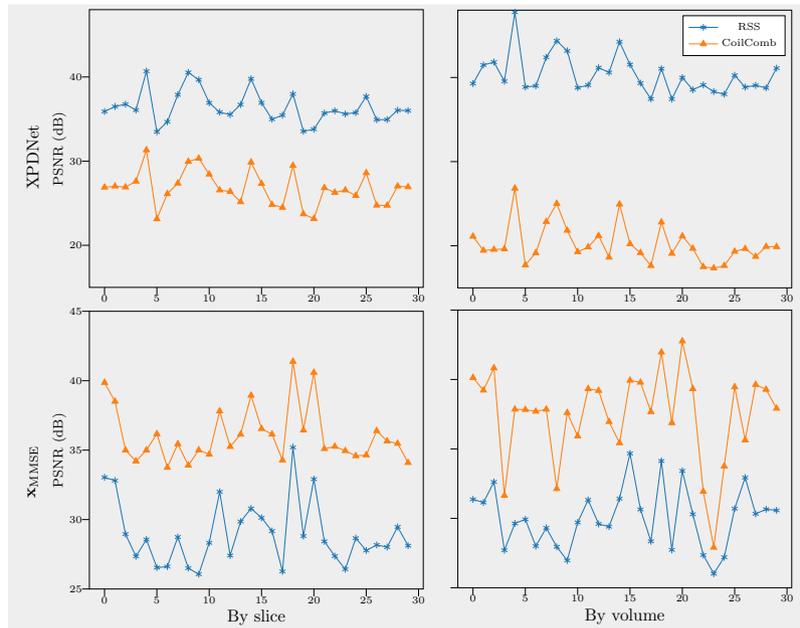

**FIGURE S3** : PSNR and SSIM metrics for different ground truths and data ranges.

## Distortion

To investigate how $\lambda$ affects the sampling process, we repeated the *more noise scales* reconstruction experiment five times with different $\lambda = \{1, 2, 3, 5, 25\}$. NET$_2$ was used to construct transition kernels and the other parameters in Algorithm 1 are $K = 2, N = 80$.

Figure S4 shows the samples that explore the distribution space with different strength of the data consistency by changing $\lambda$. Comparing samples that are reconstructed with small $\lambda$, we observed large variations between two selected samples in Figure S4a and Figure S4b and the distorted samples are far away from the truth. As $\lambda$ increases, the variations decreases shown in the variance map Figure S4c.

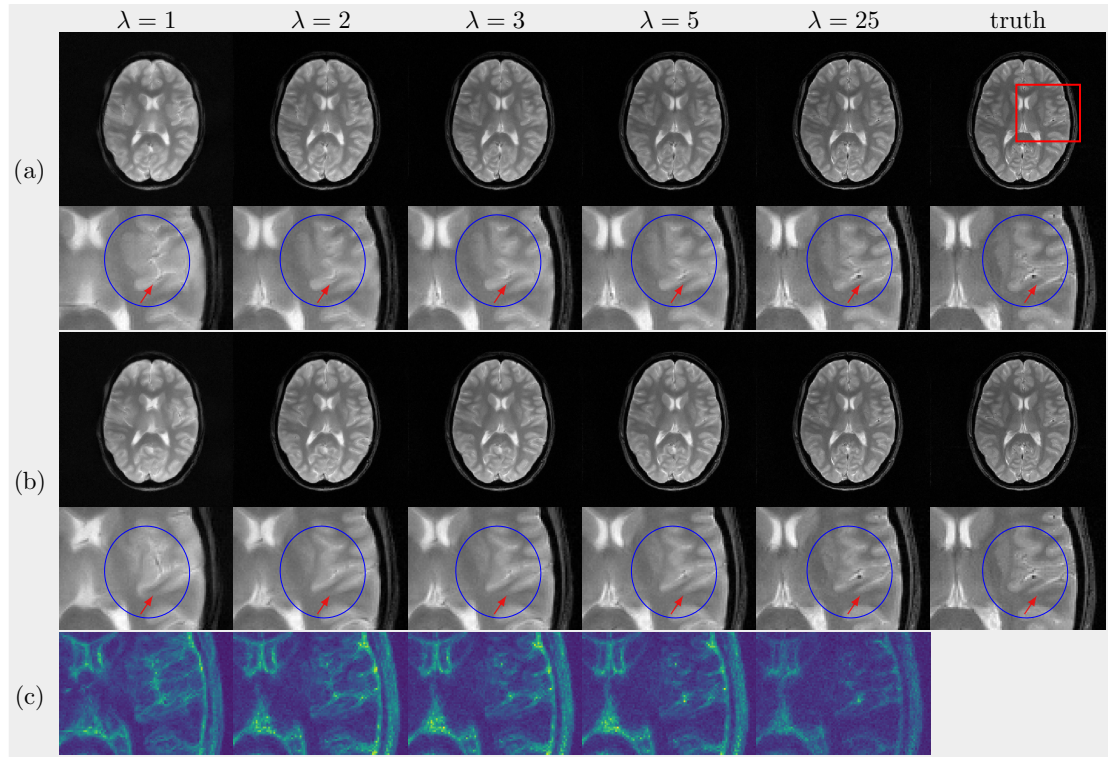

**FIGURE S4** : Samples reconstructed with different $\lambda$. Two selected samples with a particular $\lambda$ are presented in (a) and (b). The variance maps over 10 samples reconstructed with each $\lambda$ are shown in (c).

## Comparison of Prior-based Methods

Figure S5 shows that prior-based methods for reconstruction have comparable performance even for 10x undersampling pattern as quantified with the `nrmse` metric. The image reconstructed with the combined diffusion prior and autoregressive (AR) prior preserves more of the small details than the other methods.

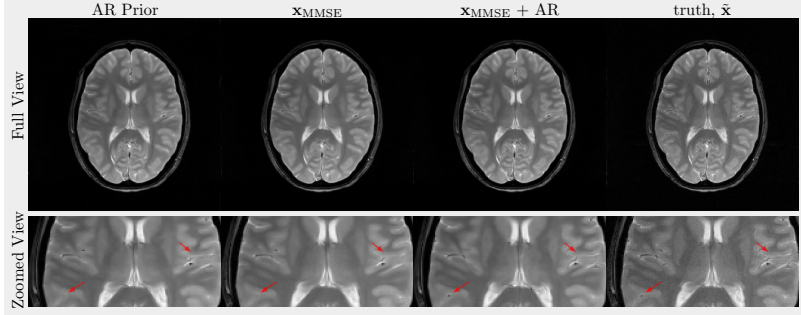

**FIGURE S5** : Reconstruction using different prior-based methods with Poisson-disc sampling with 10x undersampling in k-space. The arrows indicate small details lost in some reconstructions. The `nrmse` metrics are AR: 0.0582, $x_{MMSE}$: 0.0577, $x_{MMSE}$+AR: 0.0559. The PSNRs are AR: 36.21, $x_{MMSE}$: 36.27, $x_{MMSE}$+AR: 36.55 in dB. The SSIMs are AR: 0.8731, $x_{MMSE}$: 0.8516, $x_{MMSE}$+AR: 0.8354.

**TABLE S1** : Architectures of the score networks.

| $NET_{1/2}$ | $NET_3$ |
| --- | --- |
| 3x3 Conv2D, filters=64 | 3x3 Conv2D, filters=100 |
| CondResBlock$_1$, filters=64 | CondResBlock$_1$, filters=100 |
| CondResBlock$_1$, filters=64 | CondResBlock$_1$, filters=100 |
| CondResBlock$_2$, downsampling filters=128 | CondResBlock$_2$, downsampling filters=200 |
| CondResBlock$_2$, filters=128 | CondResBlock$_2$, filters=200 |
| CondResBlock$_3$, downsampling, filters=128, dilation=2 | CondResBlock$_3$, downsampling, filters=200, dilation=2 |
| CondResBlock$_3$, filters=128 dilation=2 | CondResBlock$_3$, filters=200 |
| | SelfAttention, filters=200 |
| CondResBlock$_4$, downsampling, filters=128, dilation=4 | CondResBlock$_4$, downsampling, filters=200, dilation=4 |
| CondResBlock$_4$, filters=128 dilation=4 | CondResBlock$_4$, filters=200 |
| | SelfAttention, filters=200 |
| CondRefineBlock$_1$, filters=128 | CondResBlock$_5$, downsampling, filters=200, dilation=4 |
| CondRefineBlock$_2$, filters=128 | CondResBlock$_5$, filters=200 |
| CondRefineBlock$_3$, filters=64 | SelfAttention, filters=200 |
| CondRefineBlock$_4$, filters=64 | CondRefineBlock$_1$, filters=200 |
| 3x3 Conv2D, filters=2 | CondRefineBlock$_2$, filters=200 |
| ($\sim$7M trainable parameters) | CondRefineBlock$_3$, filters=200 |
| | CondRefineBlock$_4$, filters=100 |
| | CondRefineBlock$_5$, filters=100 |
| | 3x3 Conv2D, filters=2 |
| | ($\sim$27M trainable parameters) |

## Hyperparameters for Training

The hyperparameters used for training networks are listed in Table S2. The indexed noise scale is $\sigma_i = \sigma(\frac{i}{N}) = \sigma_{\min}(\frac{\sigma_{\max}}{\sigma_{\min}})^{\frac{i-1}{N-1}}$.

**TABLE S2** : Hyperparameters for training

|  | $\texttt{NET}_1$ | $\texttt{NET}_2$ | $\texttt{NET}_3$ |
|---|---|---|---|
| $\sigma_{\max}/\sqrt{2}$ | 0.3 | 0.5 | 0.5 |
| $\sigma_{\min}/\sqrt{2}$ | 0.01 | 0.01 | 0.01 |
| N | 15 | 100 | 100 |
| batch size | 10 | 5 | 5 |
| embedding size | - | 128 | 200 |
| Fourier scale | - | 16 | 16 |
| learning rate | 0.0001 | 0.0001 | 0.0001 |
| activation | elu | elu | elu |
| optimizer | adam | adam | adam |

## Performance on Open Dataset

**TABLE S3** : Average PSNR (dB) and SSIM(%) for test subjects

| Sampling rate | $\ell_1$-ESPIRiT | | | $\mathbf{x}_{\mathrm{MMSE}}$ | | |
|---|---|---|---|---|---|---|
| *PSNR* | $sub_1$ | $sub_2$ | $sub_3$ | $sub_1$ | $sub_2$ | $sub_3$ |
| 10% | $32.27 \pm 1.03$ | $34.67 \pm 1.77$ | $34.03 \pm 2.98$ | $34.07 \pm 0.91$ | $36.12 \pm 1.85$ | $35.65 \pm 3.37$ |
| 20% | $33.69 \pm 0.86$ | $35.60 \pm 2.18$ | $35.17 \pm 3.41$ | $35.04 \pm 0.89$ | $37.15 \pm 2.08$ | $36.58 \pm 3.84$ |
| *SSIM* | | | | | | |
| 10% | $79.69 \pm 3.28$ | $82.74 \pm 8.57$ | $79.98 \pm 12.40$ | $84.21 \pm 3.24$ | $86.06 \pm 6.99$ | $83.24 \pm 12.07$ |
| 20% | $83.68 \pm 3.38$ | $84.60 \pm 9.20$ | $82.70 \pm 12.93$ | $86.38 \pm 3.04$ | $88.26 \pm 7.07$ | $84.85 \pm 12.99$ |



\end{document}